\documentclass[Afour,sageh,times]{sagej}
\usepackage{textcomp}
\usepackage{moreverb,url}
\usepackage{graphicx}
\usepackage{caption,subcaption}
\usepackage{pifont}
\usepackage{siunitx}
\usepackage{subcaption}

\usepackage{hyperref}
\hypersetup{colorlinks, bookmarksopen, bookmarksnumbered,
citecolor=[rgb]{0.000,0.25,0.05},
urlcolor=blue,
linkcolor=red}

\newcommand\BibTeX{{\rmfamily B\kern-.05em \textsc{i\kern-.025em b}\kern-.08em
T\kern-.1667em\lower.7ex\hbox{E}\kern-.125emX}}
\usepackage{multirow}
\usepackage{lipsum}
\usepackage{gensymb}
\usepackage[table]{xcolor} 
\usepackage{soul}

\newcommand{\xmark}{\ding{55}}%

\definecolor{top1}{RGB}{111, 170, 247}
\definecolor{top2}{RGB}{169, 204, 250}
\definecolor{top3}{RGB}{226, 238, 253}
\def\secref#1{Sec.~\ref{#1}}
\def\figref#1{Fig.~\ref{#1}}
\def\tabref#1{Tab.~\ref{#1}}
\def\eqref#1{Eq.~(\ref{#1})}

\def\volumeyear{2025}

\begin{document}

\runninghead{Tao \textit{et~al.}}

\title{The Oxford Spires Dataset: Benchmarking Large-Scale LiDAR-Visual Localisation, Reconstruction and Radiance Field Methods}

\author{Yifu Tao\affilnum{1}\thanks{denotes equal contribution}, Miguel Ángel Muñoz-Bañón\affilnum{1,2}$^{\ast}$, Lintong Zhang\affilnum{1}, Jiahao Wang\affilnum{1}, Lanke Frank Tarimo Fu\affilnum{1}, and Maurice Fallon\affilnum{1}
}

\affiliation{\affilnum{1}Oxford Robotics Inst., Dept. of Eng. Science, Univ. of Oxford, UK\\
\affilnum{2}Group of Automation, Robotics and Computer Vision (AUROVA), University of Alicante, Spain}

\corrauth{Yifu Tao,
Oxford Robotics Inst.,
Dept. of Eng. Science,
Univ. of Oxford
OX1~3PJ, UK.}

\email{oxfordspiresdataset@robots.ox.ac.uk}

\begin{abstract}
This paper introduces a large-scale multi-modal dataset captured in and around well-known landmarks in Oxford using a custom-built multi-sensor perception unit as well as a millimetre-accurate map from a Terrestrial LiDAR Scanner (TLS). The perception unit includes three synchronised global shutter colour cameras, an automotive 3D LiDAR scanner, and an inertial sensor --- all precisely calibrated. We also establish benchmarks for tasks involving localisation, reconstruction, and novel-view synthesis, which enable the evaluation of Simultaneous Localisation and Mapping (SLAM) methods, Structure-from-Motion (SfM) and Multi-view Stereo (MVS) methods as well as radiance field methods such as Neural Radiance Fields (NeRF) and 3D Gaussian Splatting. To evaluate 3D reconstruction the TLS 3D models are used as ground truth. Localisation ground truth is computed by registering the mobile LiDAR scans to the TLS 3D models. Radiance field methods are evaluated not only with poses sampled from the input trajectory, but also from viewpoints that are from trajectories which are distant from the training poses. 
Our evaluation demonstrates a key limitation of state-of-the-art radiance field methods: we show that they tend to overfit to the training poses/images and do not generalise well to out-of-sequence poses. They also underperform in 3D reconstruction compared to MVS systems using the same visual inputs. Our dataset and benchmarks are intended to facilitate better integration of radiance field methods and SLAM systems. The raw and processed data, along with software for parsing and evaluation, can be accessed at \url{https://dynamic.robots.ox.ac.uk/datasets/oxford-spires/}. 
\end{abstract}


\maketitle

\begin{figure*}[t]
\centering

\begin{subfigure}[b]{1\linewidth}
    \centering
    \includegraphics[width=1\textwidth]{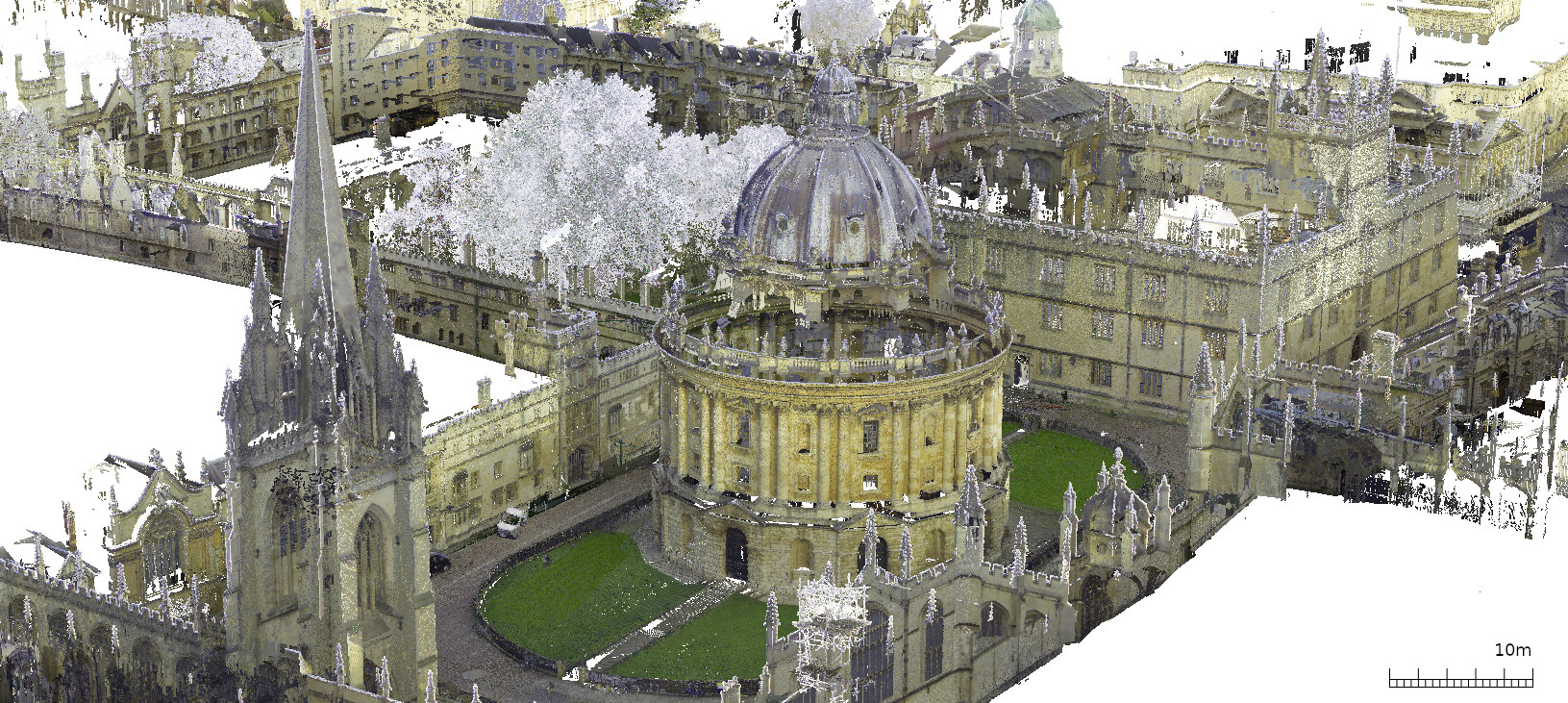}
\end{subfigure}%

\begin{subfigure}[b]{0.25\linewidth}
    \centering
    \captionsetup{justification=centering}
    \includegraphics[width=0.99\textwidth]{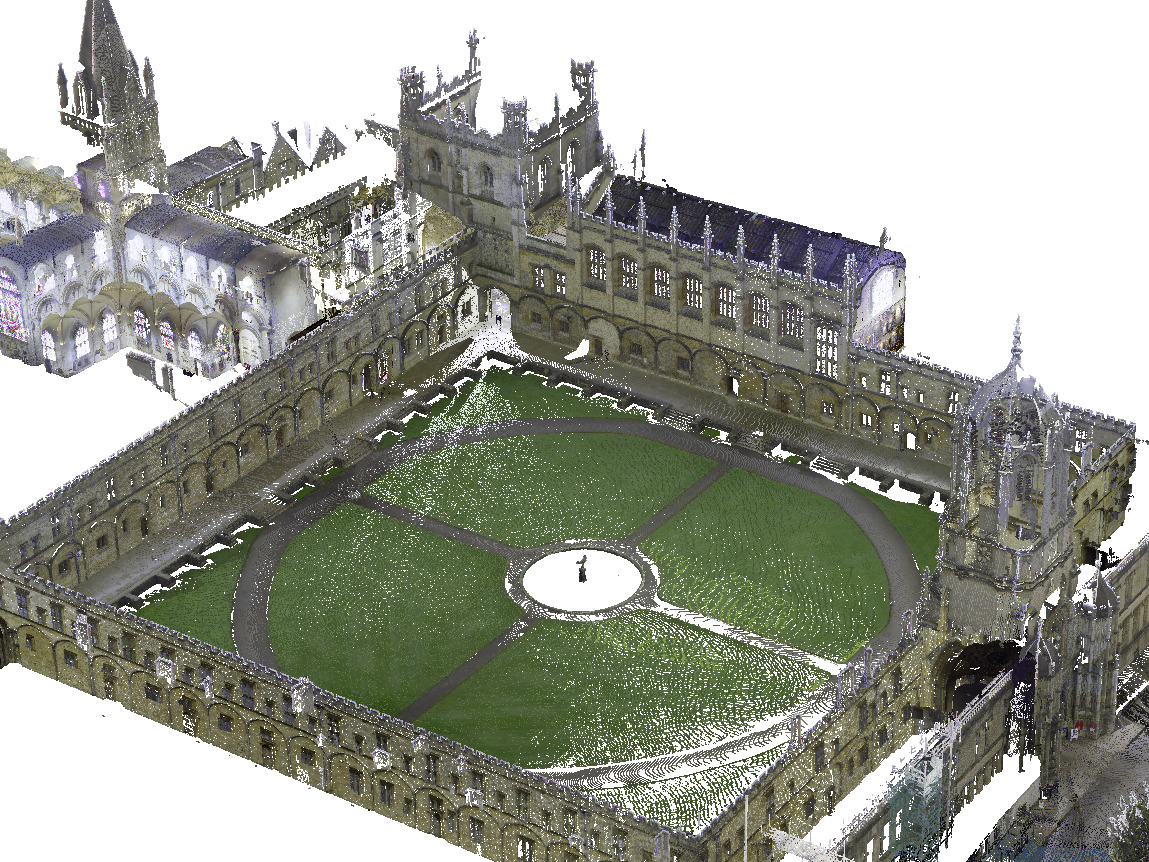}
    \includegraphics[width=0.99\textwidth]{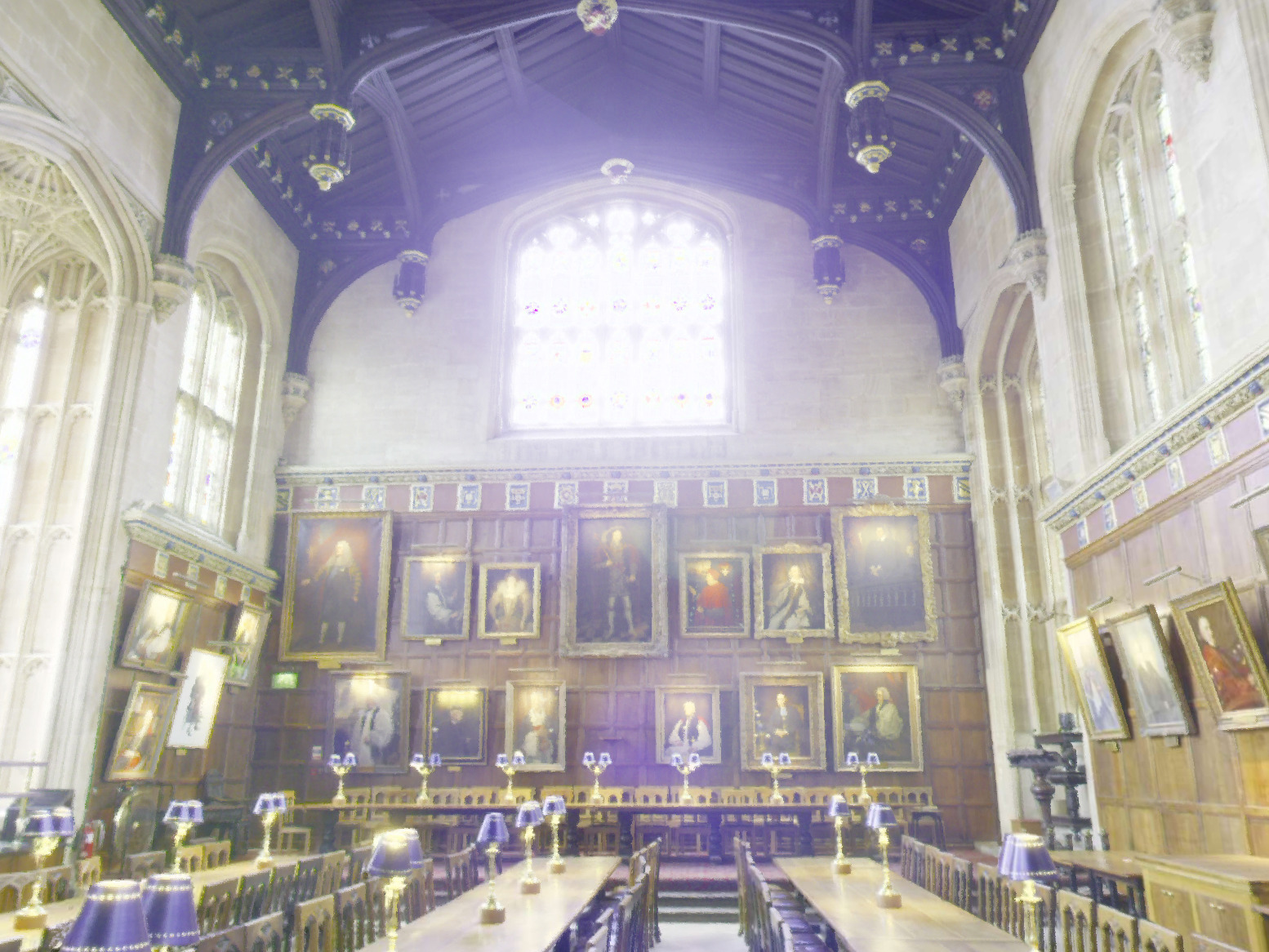}
    \caption{Christ Church College}
\end{subfigure}%
\begin{subfigure}[b]{0.25\linewidth}
    \centering
    \captionsetup{justification=centering}
    \includegraphics[width=0.99\textwidth]{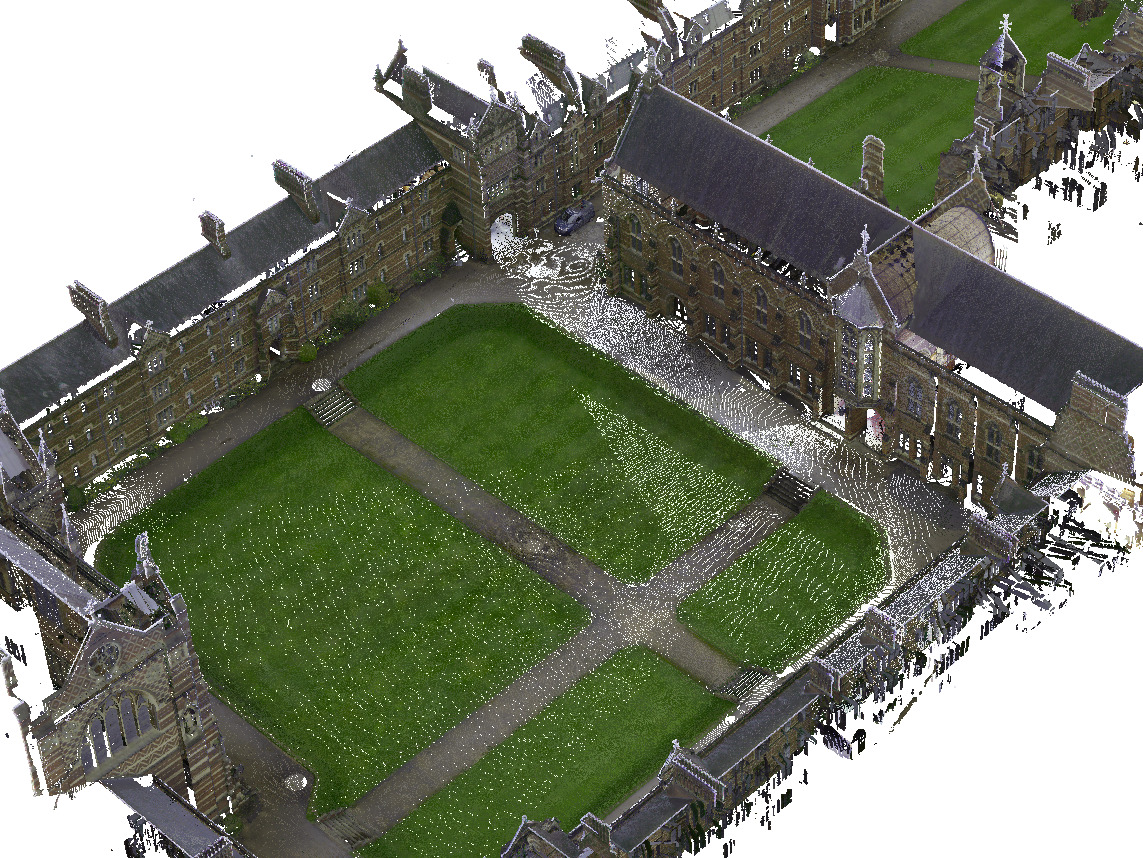}
    \includegraphics[width=0.99\textwidth]{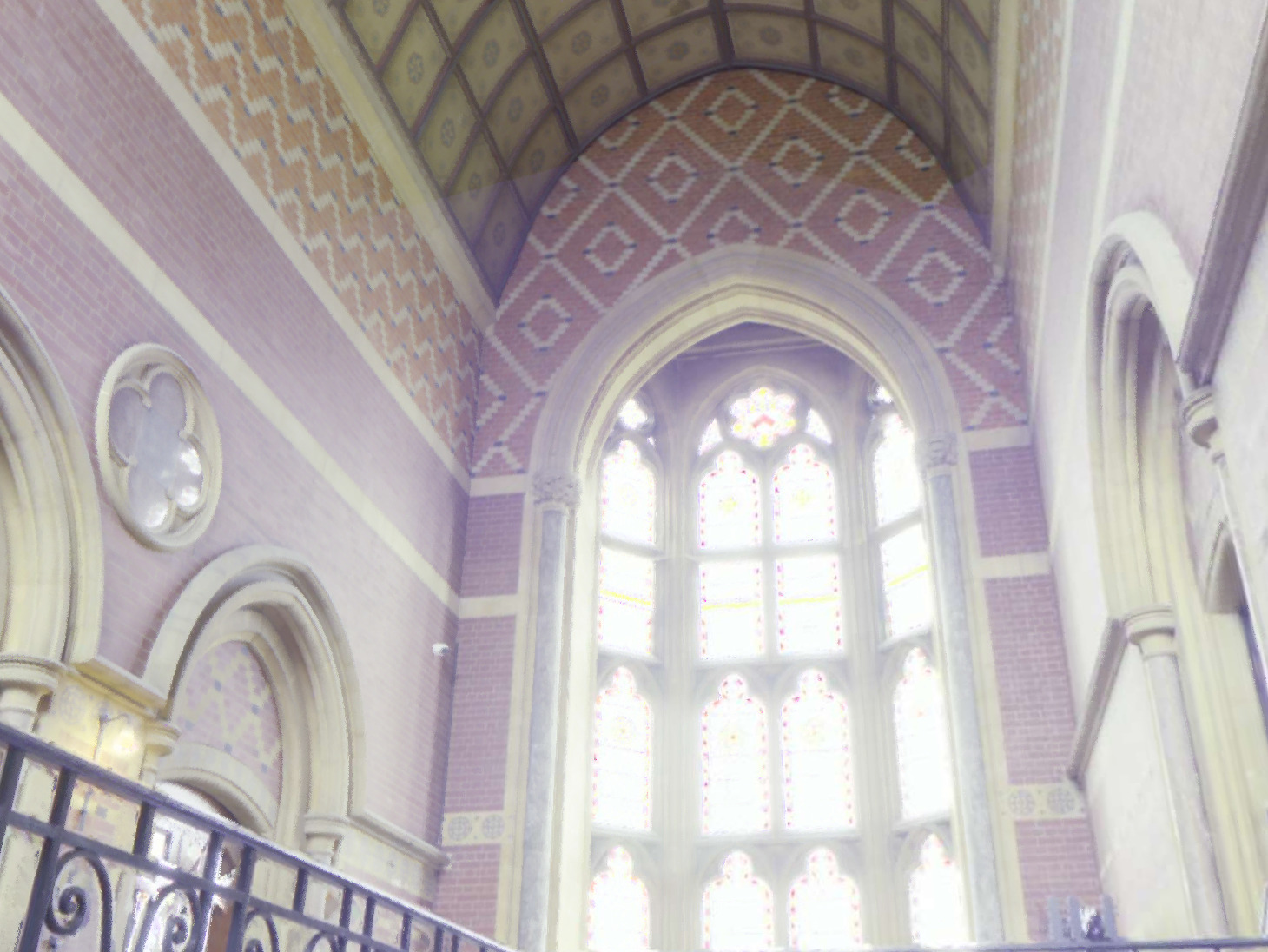}
    \caption{Keble College }
\end{subfigure}%
\begin{subfigure}[b]{0.25\linewidth}
    \centering
    \captionsetup{justification=centering}
    \includegraphics[width=0.99\textwidth]{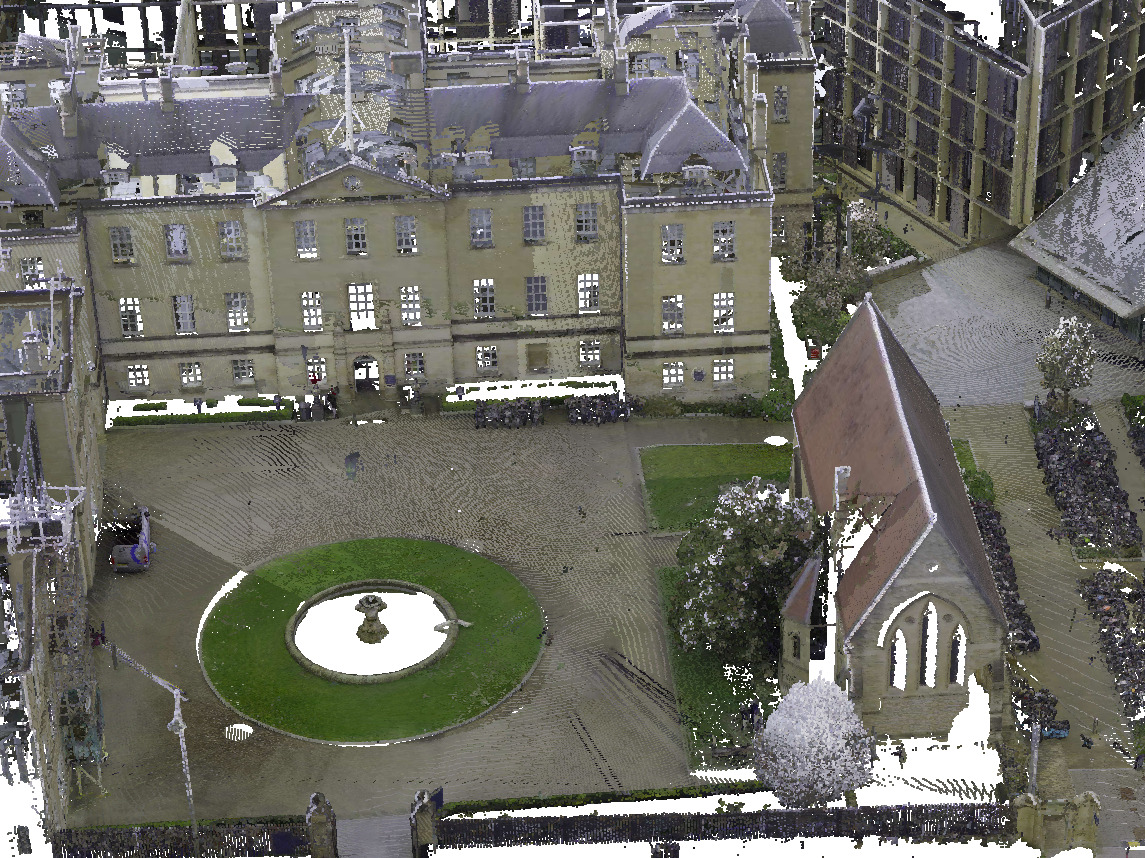}
    \includegraphics[width=0.99\textwidth]{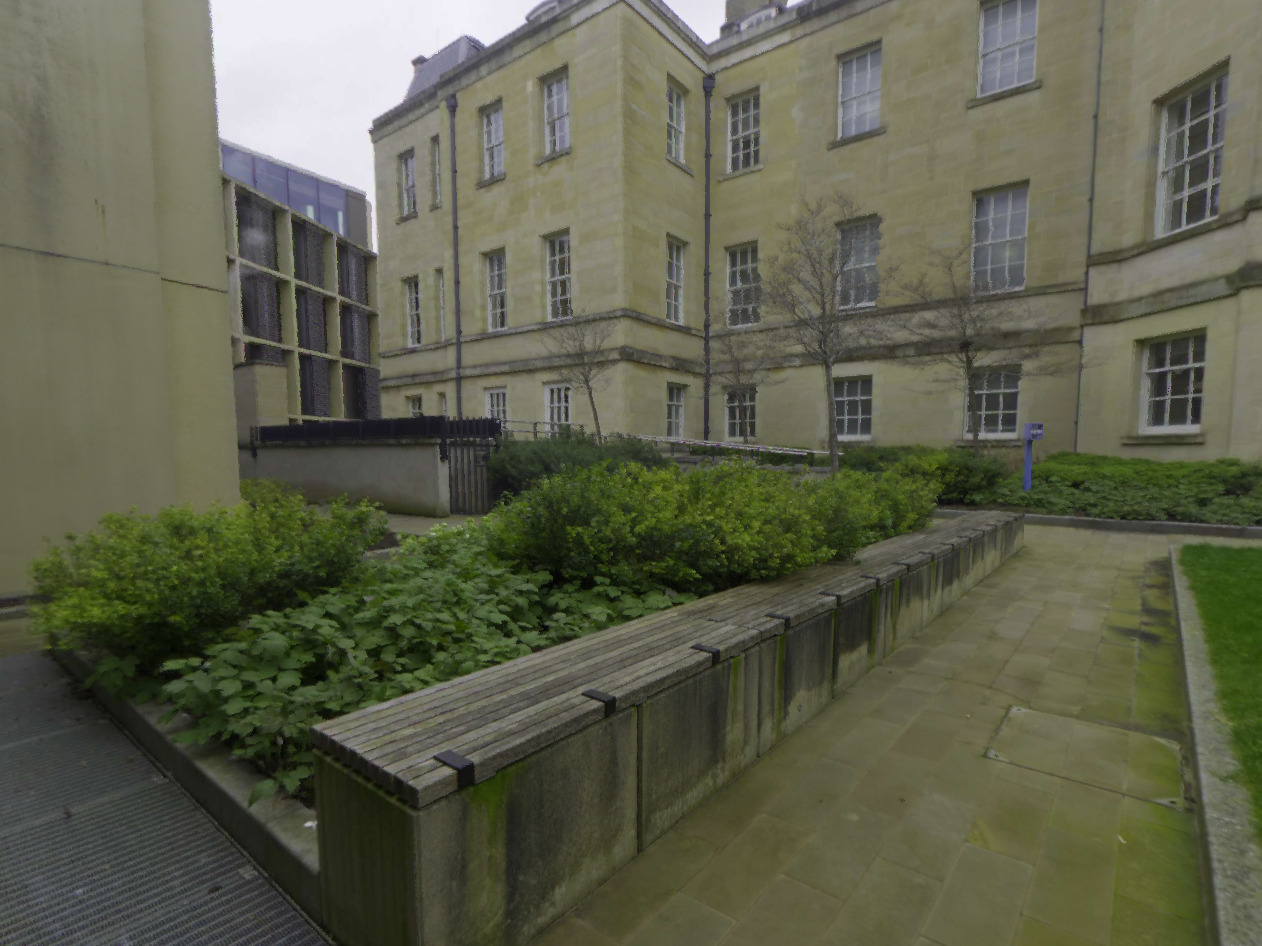}
    \caption{Radcliffe Obs. Quarter}
\end{subfigure}%
\begin{subfigure}[b]{0.25\linewidth}
    \centering
    \captionsetup{justification=centering}
    \includegraphics[width=0.99\textwidth]{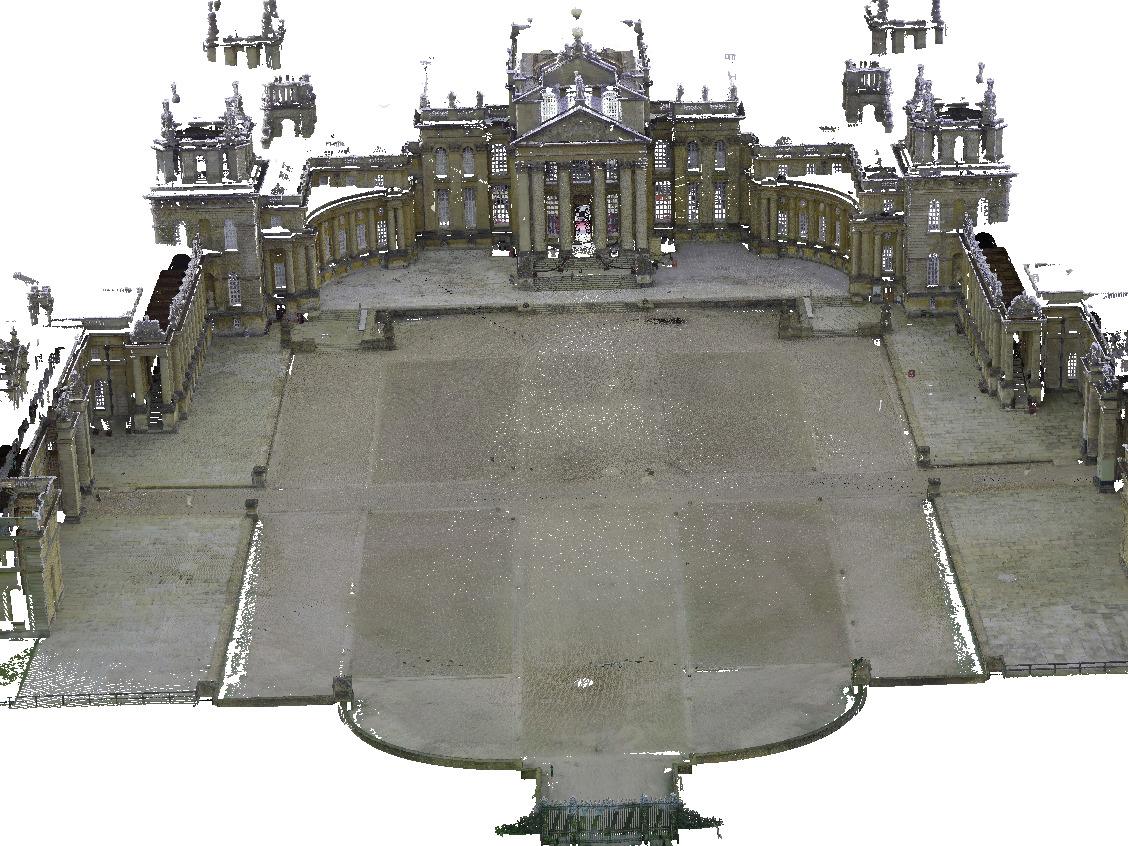}
    \includegraphics[width=0.99\textwidth]{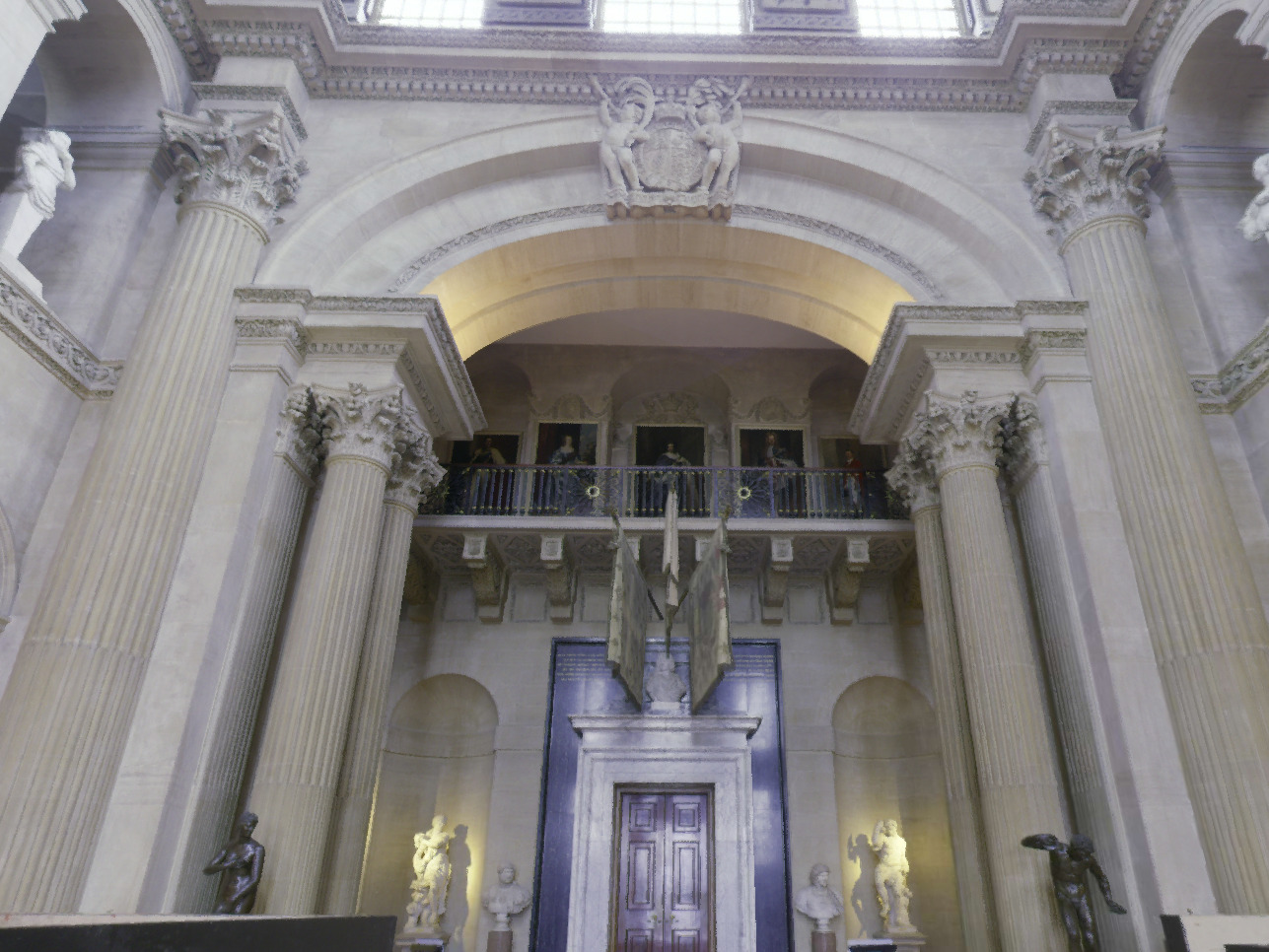}
    \caption{Blenheim Palace}
\end{subfigure}%
 
 \caption{\textit{Top}: Point cloud of the Radcliffe Camera and the Bodleian Library captured by the TLS. Note the scale bar in the bottom-right corner. \textit{Bottom}: Views of the TLS maps from other sites. Each column is a different site. The upper images show views of colour point clouds while the lower images were taken by the scanner's cameras.}
	
 \label{fig:bodleian}
\end{figure*}


\section{Introduction}

Localisation and 3D reconstruction are fundamental problems in both robotics and computer vision, with applications spanning autonomous driving, building inspection and augmented reality. There are methods that focus on localisation (e.g. visual or lidar odometry, Structure-from-Motion (SfM), place recognition and relocalisation) and others that focus on 3D reconstruction/mapping (e.g. Multi-view Stereo (MVS) and occupancy mapping). In mobile robotics, both problems can be solved concurrently by Simultaneous Localisation and Mapping (SLAM) methods, which are our primary focus in this work. For large-scale outdoor environments, cameras and LiDARs are the most commonly used exteroceptive sensor modalities for environment perception tasks. Additionally, the Inertial Measurement Unit (IMU) is a common interoceptive sensor. The two mentioned exteroceptive sensor technologies have complementary characteristics: LiDAR captures long-range depth measurements that are accurate but sparse, while camera images capture texture with higher resolution.

The evaluation of the outdoor SLAM systems has primarily focused on localisation accuracy, conversely quantitative evaluation of the 3D reconstruction quality is often lacking. An important reason for this is the limited availability of high-quality ground truth. For 3D reconstruction, ground truth is typically collected using survey-grade Terrestrial LiDAR Scanners (TLS) which are expensive~\citep{zhang2022hilti}. Compared to indoor scenes where MVS and RGB-D SLAM systems are often evaluated, the large-scale of outdoor environments makes TLS data collection laborious. As a result, many outdoor SLAM datasets do not include precise ground truth reconstruction from TLS and rely on ground truth trajectories from other sensors such GNSS-RTK~\citep{kitti}. 

In addition to geometric reconstruction, colour reconstruction is becoming more important with the advances of radiance field methods including Neural Radiance Fields (NeRF)~\citep{mildenhall2021nerf} and 3D Gaussian Splatting~\citep{kerbl3Dgaussians}. These methods take as input calibrated camera images and their precise 3D poses (typically estimated using SfM), and output a dense 3D field with volume density (similar to differential opacity) and view-dependent colour. The output radiance field can be used to synthesise photorealistic images using volume rendering techniques. Since radiance field methods are capable of representing complex geometry and appearance, some SLAM systems have adopted it as their underlying 3D map representation~\citep{sucar2021imap,zhu2022nice}. 

Despite the rapid development of radiance field methods, their use in outdoor mobile robot perception has been less well explored. Radiance field methods are often evaluated by the quality of images rendered using datasets where the image set points at a single object observed in controlled lighting conditions, and often indoors. For a mobile robot operating in an outdoor environment, the trajectory is usually not object-centric, and viewpoints are relatively sparse compared to the size of the scene. Inferring 3D structure from monocular images alone is more challenging if provided with fewer viewpoint constraints, and this can lead to artefacts (e.g. the elongated Gaussians along the viewing direction mentioned in \cite{matsuki2024gaussianslam}) that are not noticeable if only evaluated from nearby poses. In addition, radiance field methods can also generate ``floater'' artefacts to overfit per-frame lighting conditions (as discussed in \cite{tancik2023nerfstudio}) and the texture of the sky. Both are common challenges in outdoor environments. Such artefacts can lead to inferior 3D reconstruction and poorer photo-realistic rendering from a pose that is far from the training sequence. To develop radiance field methods that can be integrated with outdoor SLAM systems, it is crucial to have a dataset with colour images, LiDAR and accurate ground truth trajectory and reconstruction.

In this work, we introduce the Oxford Spires Dataset, a large-scale dataset collected across six historical landmarks in Oxford, UK, covering more than \SI{20000}{\meter^2} per site. The total area recorded in this dataset is more than \SI{125000}{\meter^2}, about the size of a small town. It provides high-resolution RGB image streams from three cameras, 3D wide Field-of-View LiDAR data, and inertial data from a mobile handheld device. It is accompanied by millimetre-accurate reference scans which serve as the reference ground truth 3D model for reconstruction systems. We also use it to determine the ground truth trajectories of the handheld device. The setup of three colour cameras facing forward, left and to the right is a particularly novel characteristic as similar datasets contain TLS-based ground truth, LiDAR scans but only forward-facing colour camera(s)~\citep{wei2024fusionportablev2,nguyen2024mcd,liu2023botanicgarden}. The side-facing cameras provide increased Field-of-View that not only increases the texture mapping coverage but also provides view constraints which are crucial for vision-only systems to infer the 3D structure. The three-camera setup also makes textured mapping more tractable from a simple linear pass through an environment, rather than requiring exhaustive scanning. This makes our dataset suitable for evaluating radiance field methods in outdoor mobile robotics and reality capture contexts. Leveraging this rich combination of sensor data, we also introduce three benchmarks for localisation, 3D reconstruction, and novel-view synthesis. We use the benchmark to evaluate state-of-the-art SLAM systems, SfM-MVS systems and radiance field systems. In particular, the novel-view synthesis benchmark features test data not only sampled from a single reference trajectory, but also from other sequences where the device travelled in an opposing direction and along a trajectory far from the reference. The evaluation results of state-of-the-art radiance field methods highlight the problem of overfitting to the training data and an inability to generalise to distant viewpoints. Our dataset opens up new research avenues in this space. We release the raw sensor data as well as processed data including example outputs from a LiDAR SLAM system (such as motion-undistorted point clouds) and a SfM system (which can be used for MVS, NeRF and 3D Gaussian Splatting), as well as ground truth trajectories and reconstruction. Software for parsing the data and evaluating the systems presented in the three benchmarks is also made available. 

In summary, our main contributions are as follows:
\begin{itemize}
    \item A large-scale outdoor dataset collected at six historical sites, covering an average area of about  \SI{10000}{\meter^2} each. In total, 24 sequences were recorded, with the average distance travelled in each sequence exceeding 400 metres. 
    \item The dataset is collected with a sensor suite comprising three 1.6 megapixel global shutter fisheye RGB cameras, a wide Field-of-View 64-beam 3D LiDAR, and inertial data, paired with millimetre-accurate reference 3D models captured using a TLS.
    \item We provide precise calibrations for the synchronised sensors --- including 3 fisheye cameras, IMU and the 104$\degree$ FoV LiDAR.
    \item Three benchmarks for localisation, reconstruction and novel-view synthesis with ground truth generated using the 3D models from the TLS. In this paper, we evaluated state-of-the-art SLAM, SfM, MVS, NeRF and 3D Gaussian Splatting methods for each.
    \item Evaluation software is released for using the dataset and benchmarking methods. 
\end{itemize}

\begin{table*}[t]
\setlength{\tabcolsep}{1.7pt}
\scriptsize	
\caption{Summary of related datasets for testing localisation and reconstruction methods. Oxford Spires is a large-scale outdoor SLAM dataset with three colour camera images, LiDAR as well as ground truth trajectories and 3D models. It can be used to evaluate tasks including localisation, reconstruction and novel-view synthesis. Features in other related datasets that are not suitable to our target domain (outdoor SLAM with colour reconstruction) are coloured with a \colorbox{gray!50}{grey background}. These features include indoor scenes, greyscale camera images, short range ($<$10 m) depth sensing, and imprecise or missing ground truth 3D models.}
\centering
\begin{tabular}{m{4.5cm} l| c c c|c|c c}
\toprule
\multirow{2}{*}{Dataset}  & \multirow{2}{*}{Scene} & \multicolumn{3}{c|}{Camera} &\multirow{2}{*}{Depth Sensor} & \multicolumn{2}{c}{Ground Truth}\\
&&Colour & Shutter & Resolution&&Trajectory&3D Model\\
\midrule
New College~\citep{smith2009new}&Outdoor&RGB/Grey&-&0.2 MP&2$\times$ LM2 291-S14&\cellcolor{gray!50}-&\cellcolor{gray!50}-\\
TUM-RGBD~\citep{sturm2012slambenchmark} &\cellcolor{gray!50}Indoor&RGB&Rolling&0.3 MP&\cellcolor{gray!50}MS Kinect RGB-D&MoCap&\cellcolor{gray!50}-\\
KITTI~\citep{kitti} &Outdoor&RGB/Grey&Global&1.4 MP&Velodyne HDL-64E&GPS RTK&\cellcolor{gray!50}-\\
NCLT~\citep{nclt}& Outdoor\&Indoor  &RGB&Global&1.9 MP&Velodyne HDL-32E&GPS RTK&\cellcolor{gray!50}-\\
EuROC~\citep{Burri2016euroc}&\cellcolor{gray!50}Indoor&\cellcolor{gray!50}Greyscale&Global&0.4 MP&\cellcolor{gray!50}-&Leica\&Vicon& Leica MS50 \\
DTU~\citep{aanaes2016large_dtu} &\cellcolor{gray!50}Indoor&RGB&-&1.9 MP& \cellcolor{gray!50}- &Robot Arm&Structured light scanner\\
ScanNet~\citep{dai2017scannet}&\cellcolor{gray!50}Indoor&RGB&Rolling&1.3 MP&\cellcolor{gray!50}Structure Sensor RGB-D&RGB-D SLAM&\cellcolor{gray!50}RGB-D SLAM\\
ETH3D~\citep{schops2017eth3d}&Outdoor\&Indoor &RGB/Grey&Global&0.4/24 MP& \cellcolor{gray!50}- &COLMAP+ICP&FARO Focus X330\\
Tanks and Temples~\citep{knapitsch2017tanks}&Outdoor\&Indoor&RGB&Rolling&8 MP& \cellcolor{gray!50}- &Mutual Information& FARO Focus X330\\
Complex Urban~\citep{jeong2019complex}& Outdoor  & RGB & Global & 0.7 MP & Velodyne VLP-16 & GPS+LiDAR SLAM&\cellcolor{gray!50}-\\
WoodScape~\citep{yogamani2019woodscape}&Outdoor&RGB&Rolling&1 MP&Velodyne HDL-64E&GNSS-IMU&\cellcolor{gray!50}-\\
Newer College~\citep{ramezani2020newer}&Outdoor & \cellcolor{gray!50}Greyscale&Global&0.4 MP&Ouster OS1-64&ICP&Leica BLK360\\
Hilti-21~\citep{helmberger2022hilti}&Outdoor\&Indoor&\cellcolor{gray!50}Greyscale&Global&1.3MP&Ouster OS0-64&MoCap/Total station&\cellcolor{gray!50}-\\
Hilti-22~\citep{zhang2022hilti} &Outdoor\&Indoor & \cellcolor{gray!50}Greyscale &Global&0.4 MP& Hesai XT32 &ICP+Reference Target &Z+F Imager 5016\\
LaMAR~\citep{sarlin2022lamar}&Outdoor\&Indoor&RGB/Grey &Both&2.0 MP& \cellcolor{gray!50}HoloLens2/iPhone RGB-D&  LiDAR SLAM+SfM&\cellcolor{gray!50} LiDAR SLAM+SfM\\
WHU-Helmet~\citep{li2023whu}&Outdoor\&Indoor& RGB & - & - & Livox Avia &  LiDAR SLAM&\cellcolor{gray!50} LiDAR SLAM\\
RELLIS-3D~\citep{jiang2021rellis}&Outdoor& RGB & Global & 2.3 MP & VLP-16\&OS1-64  &  LiDAR SLAM&\cellcolor{gray!50} LiDAR SLAM\\
ROVER~\citep{schmidt2024rover}&Outdoor& RGB/Gray & Both & 0.7 MP & -  &  Visual SLAM&\cellcolor{gray!50} -\\
Hilti-23~\citep{nair2024hilti}&\cellcolor{gray!50}Indoor&\cellcolor{gray!50}Greyscale&Global&1.0 MP&Robosense BPearl&Reference Target & \cellcolor{gray!50}Trimble X7  \\
ScanNet++~\citep{yeshwanthliu2023scannetpp}&\cellcolor{gray!50}Indoor&RGB&Rolling&2.8 MP&\cellcolor{gray!50}iPhone RGB-D&COLMAP&FARO Focus Premium\\
Hilti-24~\citep{sun2023nothing}&\cellcolor{gray!50}Indoor&\cellcolor{gray!50}-&\cellcolor{gray!50}-&\cellcolor{gray!50}-&\cellcolor{gray!50}Matterport RGB-D&ICP&\cellcolor{gray!50}Matterport RGB-D\\
MARS-LVIG~\citep{li2024marslvig}&Outdoor&RGB&Global&5.0 MP&Livox Avia & GNSS RTK&\cellcolor{gray!50}LiDAR-GNSS Mapping\\
VBR~\citep{brizi2024vbr}&Outdoor\&Indoor&RGB&Global&0.9 MP&Ouster OS0-128&LiDAR-GNSS&\cellcolor{gray!50} -\\
BotanicGarden~\citep{liu2023botanicgarden}&Outdoor&RGB/Grey&Global&2.3 MP&VLP16+Livox Avia&ICP&Leica RTC360\\
FusionPortableV2~\citep{wei2024fusionportablev2}&Outdoor\&Indoor&RGB&Global&0.7 MP &Ouster OS1-128&Prism-GNSS-IMU&Leica RTC/BLK360\\
MCD~\citep{nguyen2024mcd}&Outdoor\&Indoor&RGB&Global&1.0 MP&Ouster OS1 +Livox Mid70&ICP&Survey-grade map\\
\textbf{Oxford Spires (ours)}&Outdoor\&Indoor&RGB&Global&1.6 MP&Hesai QT64 &ICP&  Leica RTC360\\
\bottomrule
\addlinespace
\end{tabular}
\label{tab:related_work}
\end{table*}
\section{Related Work}
\label{sec:related_works}
In this section, we overview related datasets that are available for evaluating localisation, 3D reconstruction and novel-view synthesis. A summary of these datasets is presented in \tabref{tab:related_work}.
\subsection{Datasets for Evaluating Localisation}
Localisation is a key task in robotics and computer vision, and is performed by methods including odometry, SLAM, SfM, place recognition or relocalisation in a prior map. In indoor environments, cameras and RGB-D sensors are commonly used. TUM RGB-D~\citep{sturm2012slambenchmark} is one of the first benchmarks which sought to evaluate localisation performance using ground truth trajectories. While for visual-inertial SLAM systems, EuROC~\citep{Burri2016euroc} and TUM VI~\citep{schubert2018tumvi} are popular datasets used in the research community. 
For outdoor environments, LiDAR is a common sensor modality and has been used in robotics datasets such as New College~\citep{smith2009new} and NCLT~\citep{nclt}. Other datasets focus on evaluating odometry and SLAM trajectories in the context of autonomous driving, including KITTI~\citep{kitti}, Complex Urban~\citep{jeong2019complex}, and WoodScape~\citep{yogamani2019woodscape}. Several datasets including RELLIS-3D~\citep{jiang2021rellis}, Botanic Garden~\citep{liu2023botanicgarden}, and WHU-Helmet~\citep{li2023whu} focus on unstructured natural environments (forest and rural areas). Robot platforms were used to collect datasets including quadrupeds~\citep{brizi2024vbr,wei2024fusionportablev2} and aerial robots~\citep{li2024marslvig}.

To evaluate the accuracy of localisation systems, a precise ground truth trajectory is essential. 
For self-driving datasets, ground truth trajectories are often obtained by fusing GNSS data with inertial and LiDAR data~\citep{kitti}. One limitation of GNSS-based ground truth is that it is not reliable in areas such as urban canyons. Motion capture systems can also be used to obtain ground truth trajectories~\citep{helmberger2022hilti}, although they are often limited to indoor environments.
In outdoor environments, Newer College~\citep{ramezani2020newer} generate a centimetre-accurate ground truth by registering LiDAR scans against an accurate prior map obtained using TLS. The approach of registering mobile lidar scans to an accurate prior map was also adopted by the authors of BotanicGarden~\citep{liu2023botanicgarden} and MCD~\citep{nguyen2024mcd}. Hilti-Oxford~\citep{zhang2022hilti} is notable in achieving millimetre accuracy ground truth for a sample set of stationary poses by using reference targets. Our work follows the approach used in Newer College~\citep{ramezani2020newer} to generate dense ground truth trajectories.

\subsection{Datasets for Evaluating 3D Reconstruction}
SLAM systems estimate both a robot/sensor trajectory and the map of their environment; however, many SLAM datasets only provide ground truth trajectories. Few datasets evaluate the accuracy of the map reconstruction, because accurate ground truth reconstructions are costly and laborious to obtain. Because of this, some SLAM datasets such as ICL-NUIM~\citep{handa2014iclnuim} create ground truth 3D models using simulation, and other datasets including Matterport3D~\citep{Matterport3D} and ScanNet~\citep{dai2017scannet} actually use the output from a RGB-D SLAM system as ground truth. Replica~\citep{replica19arxiv} provides higher-quality 3D meshes than ScanNet and Matterport 3D, and the rendered images are photo-realistic. The RGB-D SLAM ground truth approach cannot be adapted to outdoor scenes due to the short range of depth cameras. LaMAR~\citep{sarlin2022lamar} includes outdoor sequences and a ground truth 3D model from a combination of VIO, SLAM and SfM. The ground truth 3D model obtained with SLAM is generally not as accurate as what TLS can produce. Survey-grade TLS achieves millimetre-level accuracy (A comparison can be found in ScanNet++~\citep{yeshwanthliu2023scannetpp}).

Among the datasets that provide precise ground truth reconstruction (obtained from TLS), EuROC~\citep{Burri2016euroc} and ScanNet++~\citep{yeshwanthliu2023scannetpp} are captured from indoor environments, and hence LiDAR is not used. The only available outdoor SLAM datasets that include accurate ground truth 3D models (to the best of our knowledge) are Newer College~\citep{ramezani2020newer} and Hilti-Oxford-2022~\citep{zhang2022hilti}.
Both datasets use relatively low-resolution greyscale cameras, and therefore are not suitable for colour 3D reconstruction. Compared to them, our dataset provides high-resolution colour images from three cameras, and is hence suitable for not only 3D reconstruction but also novel-view synthesis. 

In the field of computer vision, datasets with accurate ground truth reconstruction exist for MVS research, but often they target small-scale indoor scenes.
Middlebury~\citep{seitz2006middlebury} was one of the early datasets with ground truth depth obtained using a structured light scanner. DTU~\citep{aanaes2016large_dtu} captured individual objects using a robotic arm in a controlled environment, with its ground truth also obtained using structured light scans. ETH3D~\citep{schops2017eth3d} provides both high-resolution images ($<$80 per sequence) recorded by a DSLR camera, and low-resolution synchronised grey-scale images ($\sim$1000 per sequence), with a ground truth 3D model obtained using a TLS. Tanks and Temples~\citep{knapitsch2017tanks} is another popular benchmark for 3D reconstruction with ground truth from TLS. It can be used to evaluate both SfM and MVS algorithms and uses a higher-quality camera for its video data.

\subsection{Datasets for Evaluating Novel-view Synthesis}
Radiance fields have emerged to the most promising representation for novel-view synthesis. The input images for radiance field methods are often co-registered using SfM methods such as COLMAP~\citep{schoenberger2016colmap}. In the original NeRF paper~\citep{mildenhall2021nerf}, both synthetic datasets and real-world image sequences from LLFF~\citep{mildenhall2019llff} were used. Subsequently, Mip-NeRF 360~\citep{barron2022mipnerf360} included object-centric framed images taken in indoor and outdoor environments, and is popular in the radiance field community. Radiance field methods are typically evaluated using test set images that are sampled from an input trajectory and excluded from training. ScanNet++~\citep{yeshwanthliu2023scannetpp} used a more challenging evaluation approach. The authors capture test images independently from the training sequence using a higher-quality DSLR camera in the indoor environment. 
In contrast, our dataset focuses on large-scale outdoor environments, and provides test images captured from sequences with distant viewpoints. In this manner, we aim to advance the generalisation capability of existing radiance field methods.


\section{Hardware}
\label{sec:hardware}
\subsection{Handheld Perception Unit}
\label{sec:hardware_perception}
\begin{figure}[t]
\centering \includegraphics[width=0.8\columnwidth]{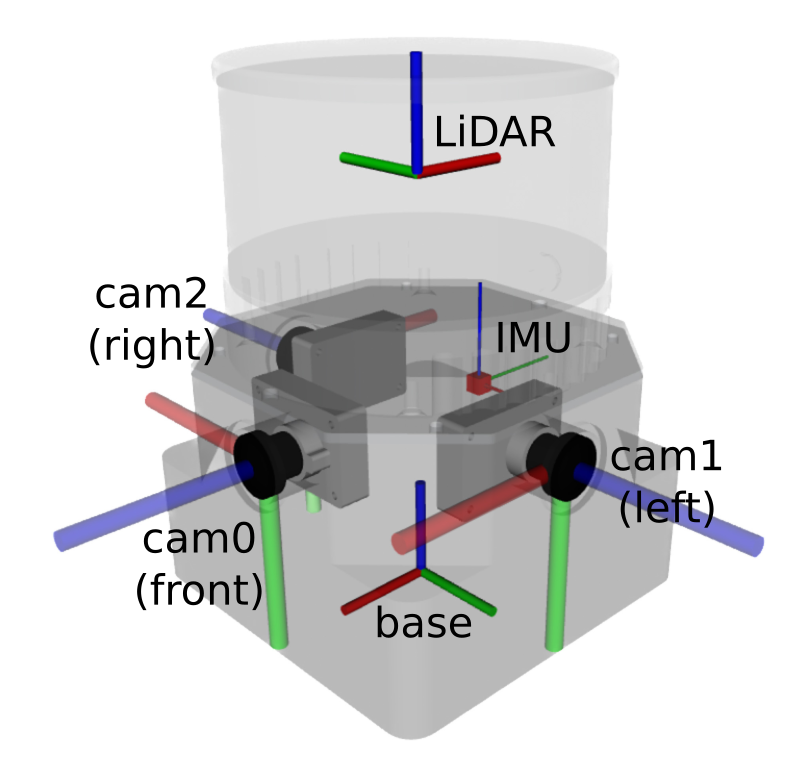}
	\caption{An isometric view of the sensor setup highlighting the coordinate frames of the cameras, the IMU and the LiDAR.}    
	\label{fig:frontier}
\end{figure}

\begin{figure}[t]
\centering \includegraphics[width=1\columnwidth]{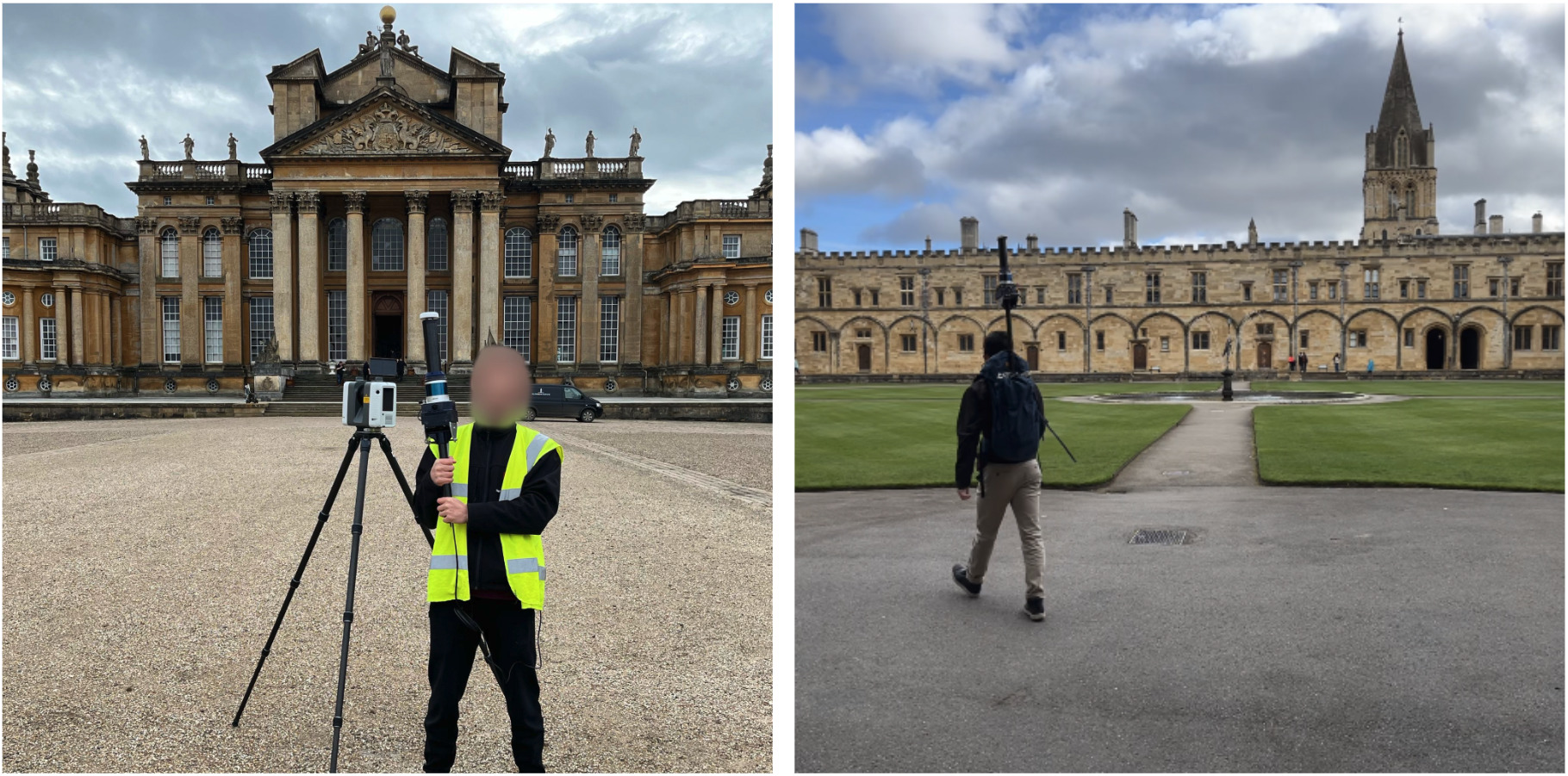}
	\caption{Leica RTC360 TLS and the Frontier device in Blenheim Palace (left) and Christ Church College (right).
    }    
	\label{fig:frontier_mounted}
\end{figure}

Our perception unit, called Frontier, has three cameras, an IMU and a LiDAR. It is shown in \figref{fig:frontier} and \figref{fig:frontier_mounted} (right). The three colour fisheye cameras face forward, left, and right from a customised Alphasense Core Development Kit from Sevensense Robotics AG. Each camera has a Field-of-View of \SI{126}{\degree}$\times$\SI{92.4}{\degree} with a resolution of 1440$\times$1080 pixels. There is a cellphone-grade IMU in the Alphasense Core Development Kit, which is hardware-synchronised with the three cameras using an FPGA from Sevensense. The three cameras have around \SI{36}{\degree} overlap, which enables multi-camera calibration mentioned in \secref{subsec:camera_calib}. The cameras operate at 20 Hz, and the IMU operates at 400 Hz. Auto-exposure is enabled for the cameras to capture indoor and outdoor scenes with different lighting conditions. A 64-channel Hesai QT64 LiDAR operating at 10 Hz was mounted on top of the cameras, with a Field-of-View of \SI{104}{\degree}, and a maximum range of \SI{60}{\meter}. The LiDAR's accuracy is $\pm$\SI{3}{cm} (typical), and the relation between reflectance and intensity is $reflectance = \alpha \times intensity \times range^2$.

Multi-sensor synchronisation is achieved both on the hardware level and software level. In terms of software synchronisation, we synchronise the device clock of Alphasense Core Development Kit (with the three cameras and IMU) and the Hesai LiDAR to the host computer on the Frontier using the Precision Time Protocol (PTP) which achieves sub-microsecond accuracy. As for hardware synchronisation, the three cameras are synchronised with the IMU within the Alphasense Core Development Kit, and we synchronise a LiDAR point cloud with the cameras using motion undistortion. While the three cameras' exposure times are not identical for one shutter cycle due to the auto-exposure, the exposure intervals of the cameras are aligned in the middle\footnote{\url{https://github.com/sevensense-robotics/core_research_manual/}}. This means that the timestamps of the mid-frame of the exposure interval of each camera are identical. This timestamp is used as the timestamp for the camera image. The Hesai QT64 LiDAR is a rolling shutter sensor, and the point cloud data is not triggered at a specific timestamp but is obtained continuously. To obtain a synchronised LiDAR point cloud for a synchronised set of three camera images, we motion-correct a LiDAR point cloud with IMU preintegration using VILENS~\citep{wisth2023vilens}. The point cloud is undistorted to the same time as the next camera image frame. More information regarding the motion undistortion can be found in \cite{wisth2021unified}. In summary, each node in the SLAM pose graph has three camera images and an undistorted LiDAR point cloud with identical timestamps, which can be used for 3D reconstruction and novel-view synthesis.

\subsection{Millimetre-accurate TLS}
\label{sec:leica}
To obtain an accurate 3D reference model to benchmark localisation and reconstruction, we used a Leica RTC360 TLS (\figref{fig:frontier_mounted}, left). It has a maximum range of 130 m and a Field-of-View of \SI{360}{\degree} $\times$ \SI{300}{\degree}. 
The final 3D point accuracy is \SI{1.9}{\milli\meter} at \SI{10}{\meter} and \SI{5.3}{\milli\meter} at \SI{40}{\meter}. The point clouds are coloured using 432 mega-pixel images captured by three cameras. Scans are registered in the field and re-optimised later using Leica's Cyclone REGISTER 360 Plus software. The average cloud-to-cloud error in our sites ranges from 3-7 \SI{}{\milli\meter}. After merging all the scans, the resultant point cloud is very large (10GB), so for ease of use we downsampled the TLS scan to 1cm resolution. Nonetheless, we provide the original raw TLS scans in our dataset.

\section{Calibration}
\label{sec:calibration}
\subsection{Multi-Camera Intrinsic and Extrinsic Parameters Calibration}\label{subsec:camera_calib}
Multi-camera sensor fusion requires an accurate camera projection modelling as well as accurate inter-camera extrinsic transforms. Given the wide Field-of-View and strong distortion of the fisheye camera lenses, we employ the equidistant distortion model~\citep{equidistant2006distortion}. Adjacent cameras in our setup share overlapping view frustums (approximately 36\degree~horizontally), enabling the extrinsic calibration between the cameras via co-detection of known calibration target features. We calibrate both the intrinsic and extrinsic parameters of the three cameras using the Kalibr open-source camera calibration toolbox~\citep{kalibr}. The resultant calibration achieves sub-pixel reprojection error, as summarised in~\tabref{tab:calibration_residuals}. We provide the camera calibration sequences with this dataset to facilitate experimentation with alternative calibration methods.

\begin{figure}[t]
	\centering
	\includegraphics[width=\linewidth]{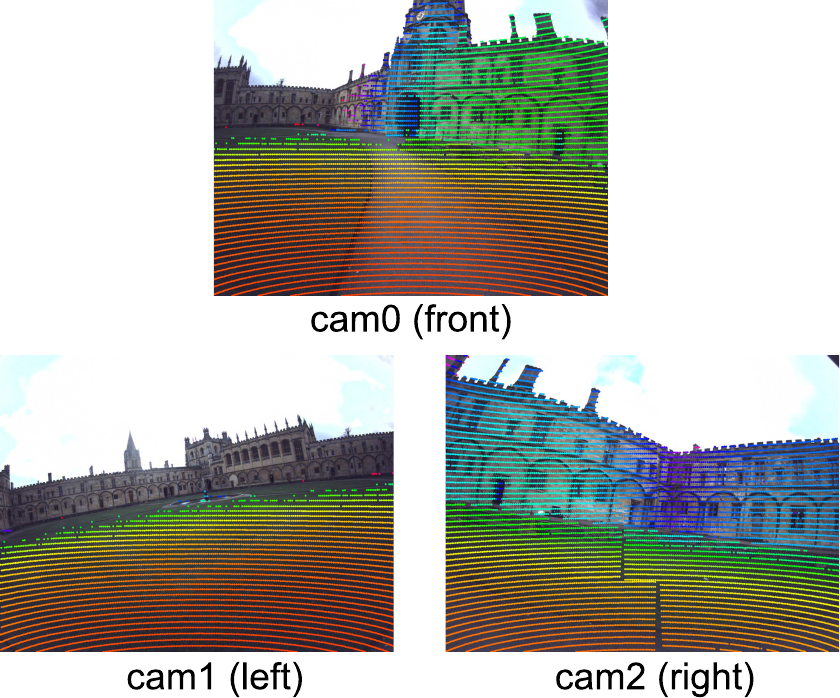}
	\caption{LiDAR point clouds overlaid on the camera images. This demonstrates the quality of camera intrinsics calibration and camera-LiDAR extrinsics calibration. In the left camera, the regions of the building without LiDAR points are due to the LiDAR's limited sensing range. Note that the motion undistortion process produces the jagged discontinuity in the LiDAR beam pattern in the right camera.}
	\label{fig:overlay}
\end{figure}

\subsection{IMU Calibration}
To facilitate visual and lidar odometry using the IMU, it is crucial to appropriately model the noise parameters of the IMU. For this, we measured the Allan variance parameters\footnote{\url{https://github.com/ori-drs/allan_variance_ros}.} of the IMU accelerometer and gyroscope using an eight-hour data sequence.

\subsection{Camera-IMU Extrinsic Calibration}
With these accurate camera intrinsic parameters and IMU noise process parameters, we then perform camera-to-IMU extrinsic calibration individually for each of the cameras using Kalibr~\citep{kalibr}. We cross-validated the consistency of the camera-IMU calibration by measuring the variation in the estimated coordinates of the IMU, using the individual camera-IMU extrinsic parameters and the camera-camera extrinsic parameters.

\subsection{Camera-LiDAR Extrinsic Calibration}
\begin{figure}[t]
    \centering
    \begin{subfigure}{\linewidth}
    \includegraphics[width=\linewidth]{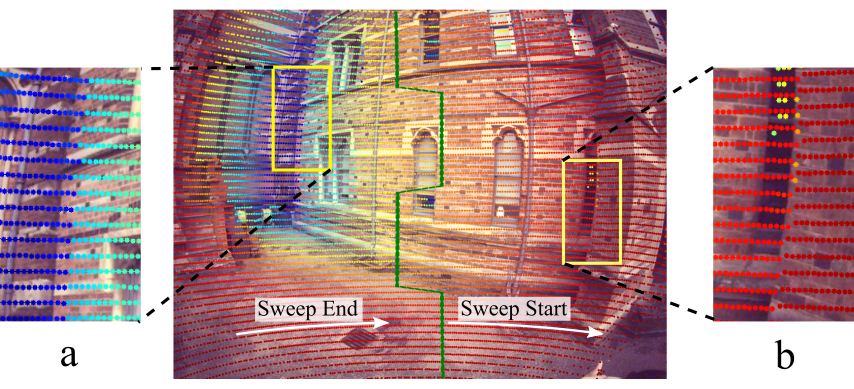}
    \caption{LiDAR-camera overlay - without motion undistortion}
    \end{subfigure}
    \begin{subfigure}{\linewidth}
    \includegraphics[width=\linewidth]{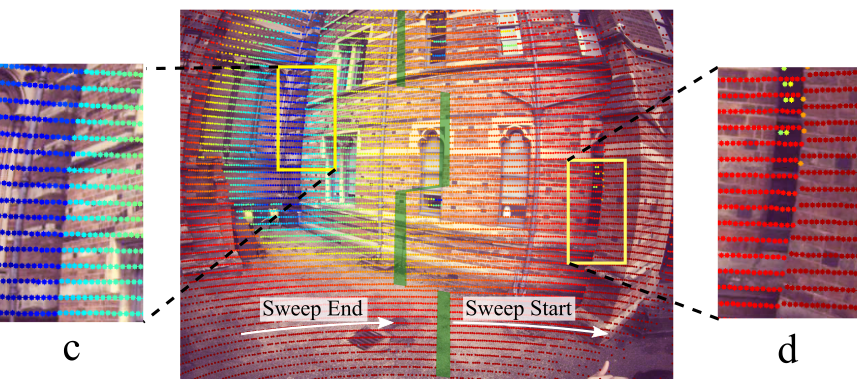}
    \caption{LiDAR-camera overlay - with motion undistortion}
    \end{subfigure}
    \caption{This image shows LiDAR overlay on top of the right-facing camera (cam2) demonstrating the effects of accurate LiDAR ego-motion undistortion. The motion here is approximately~\SI{0.9}{m/s} and~\SI{30}{\degree/s}, and the scene depth ranges between~\SI{1}{\meter} and~\SI{10}{\meter}. The right camera overlaps the start and end of the LiDAR sweep (the seam shown in green). The top figure shows the unprocessed raw point cloud overlay. The LiDAR points are slightly misaligned closer to the start of the sweep (\textbf{b}), and significantly misaligned at the end of the LiDAR sweep (\textbf{a}). With IMU-only point cloud undistortion, we achieve consistent overlays (bottom figure) (\textbf{c}) and (\textbf{d}), demonstrating the accuracy of the multi-modal spatiotemporal calibration.}
    \label{fig:lidar_undistortion_overlay}
\end{figure}

Camera-LiDAR extrinsic parameters are calibrated in a bundled fashion with the inter-camera extrinsic parameters from~\ref{subsec:camera_calib} held constant; a single $\mathit{SE}(3)$ transform between the bundle of cameras and the LiDAR is calibrated. We perform this calibration using DiffCal~\citep{diffcal}. This method uses a differentiable representation of the checker pattern to align the point intensities observed by the LiDAR directly with the camera-detected checkerboard pattern. In \figref{fig:overlay}, We present an example of the LiDAR point clouds overlaid on the camera images using the described calibration. Furthermore, \figref{fig:lidar_undistortion_overlay} illustrates the accuracy of the IMU-aided LiDAR point cloud motion undistortion by comparing the projection of raw and motion-undistorted LiDAR points onto the right-facing camera.

\begin{table}[h]
    \centering
    \begin{tabular}{lccc}
        \toprule
        \textbf{Calibration} & \textbf{Mean} (px) & \textbf{Median}  (px) & \boldmath{$\sigma$}  (px) \\ 
        \midrule
        \multicolumn{4}{l}{\textbf{Camera/Camera}} \\ 
        cam0 (front) & 0.22 & 0.21 & 0.14  \\ 
        cam1 (left)  & 0.23  & 0.22  & 0.15 \\ 
        cam2 (right) & 0.22  & 0.21  & 0.14  \\ 
        \midrule
        \multicolumn{4}{l}{\textbf{Camera/IMU}} \\ 
        cam0 (front) & 0.25 & 0.23 & 0.15  \\ 
        \bottomrule
    \end{tabular}
    \caption{Mean, median, and standard deviation of reprojection residuals for different calibration types.}
    \label{tab:calibration_residuals}
\end{table}


\begin{figure*}[t]
	\centering
	\includegraphics[width=\linewidth]{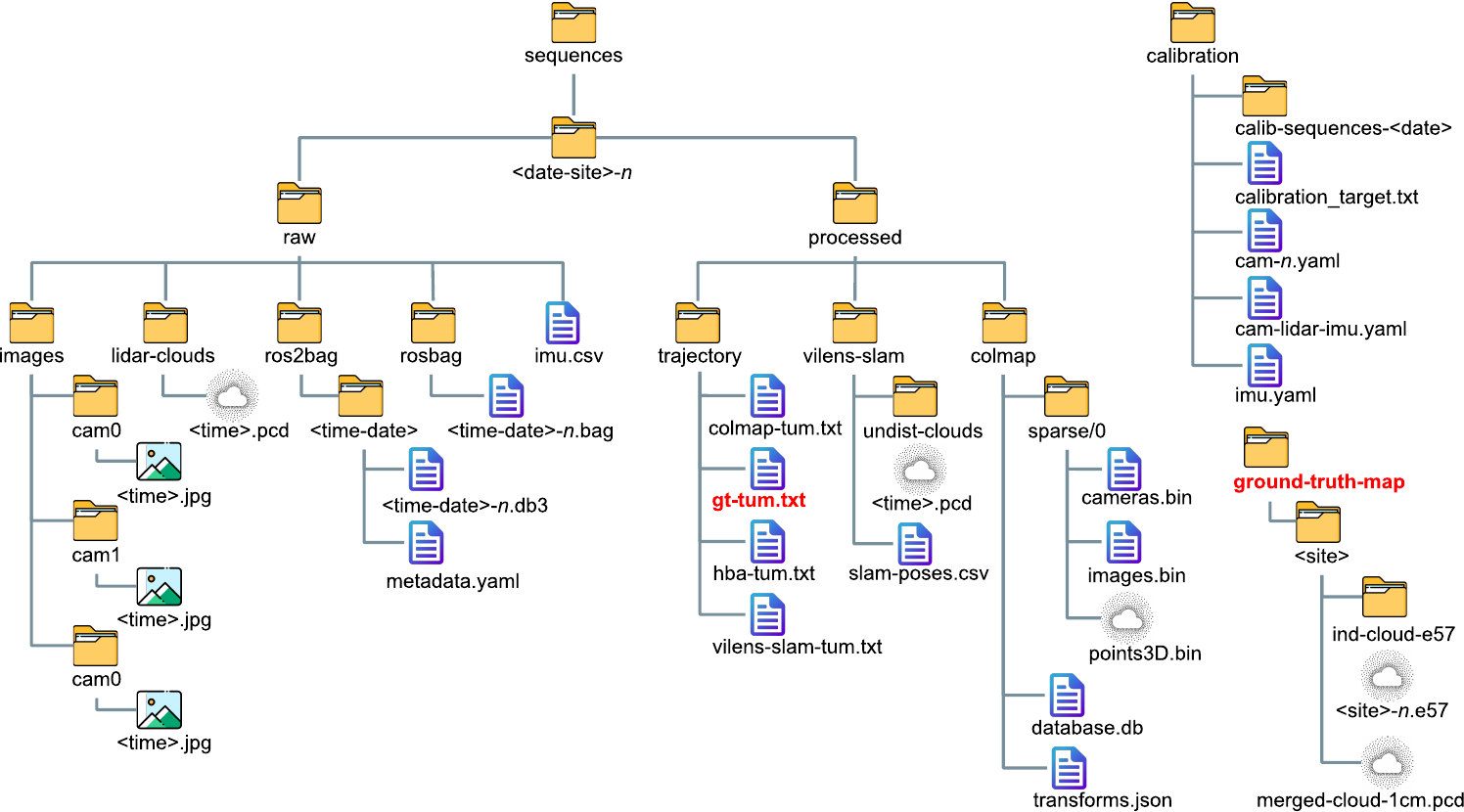}
	\caption{File structure of the Oxford Spires dataset: For \textit{each sequence}, we provide the raw images and LiDAR point clouds, ground truth trajectory, LiDAR SLAM trajectory (including undistorted point clouds synchronised to images), and COLMAP trajectory. For \textit{each site}, we provide TLS clouds as reconstruction ground truth. The ground truth systems are highlighted in red. We also provide calibration files for the camera intrinsics.}
	\label{fig:file_structure}
\end{figure*}

\section{Dataset}
\label{sec:dataset}
\subsection{Data Format}
\label{sec:data_format}
The Oxford Spires Dataset consists of data collected in six sites in Oxford, UK, with multiple sequences taken at each site (\secref{sec:sequences}). The data is originally collected as rosbags\footnote{We provide rosbags in ROS1 and ROS2 format}. We also provide raw sensor data (as individual files) as well as processed data. The processed data includes outputs from an example LiDAR SLAM system and a SfM system. Finally, we also provide the ground truth trajectories and reconstruction. 

The following sections describe the raw data formats and the folder system which we provide for easy use of the data outside of ROS (\figref{fig:file_structure}).

\subsubsection{Raw - Camera Images:}
\label{sec:raw_image}
The 20 Hz raw colour fisheye image streams from the three cameras of the Frontier are debayered and stored as 8-bit JPEG images. The three cameras are hardware-synchronised with each other, and hence the image triplets have the same timestamps. The images are stored as \texttt{<time>.jpg} under each camera folder, namely \texttt{cam0}, \texttt{cam1} and \texttt{cam2}, which correspond to the camera facing forward, left and right respectively. We debayer the raw image with bilinear interpolation, and white-balance the debayered image using the convolutional colour constancy method~\citep{barron2017fast}. We provide tools to white-balance\footnote{Tool based on \url{https://github.com/leggedrobotics/raw\_image\_pipeline}} the debayered images.

\subsubsection{Raw - 3D LiDAR Point Clouds:}
3D point clouds were collected using a Hesai QT64 LiDAR at 10Hz, and stored as \texttt{<time>.pcd}. Note that the unprocessed point clouds are raw measurements from the continuously scanning LiDAR, and the timestamp is the start time for each sweep. A subset of the LiDAR point clouds that were output by an example LiDAR SLAM system are also provided and described in \secref{sec:slam_outputs}.

\subsubsection{Raw - IMU Measurements:}
The linear acceleration and angular velocity measurements from the IMU are stored in \texttt{imu.csv}. Each row has the format of timestamp (seconds, nano-seconds), acceleration (x,y,z) and angular-velocity (x,y,z).

\subsubsection{Processed - VILENS-SLAM Outputs:}
\label{sec:slam_outputs}
 We provide the estimated trajectory and the motion undistorted point clouds output by LiDAR-inertial SLAM (VILENS-SLAM~\citep{wisth2023vilens,ramezani2020slam}). The trajectory is saved as \texttt{slam\_poses.csv} with a $\mathit{SE}(3)$ pose estimate consisting of position (x,y,z) and quaternion (x,y,z,w) for each timestamp (seconds, nano-seconds). This data format can be directly used to test 3D reconstruction.

\begin{figure}[t]
\centering
\begin{subfigure}[b]{1.0\linewidth}
    \centering
    \captionsetup{justification=centering}
    \includegraphics[width=0.99\textwidth]{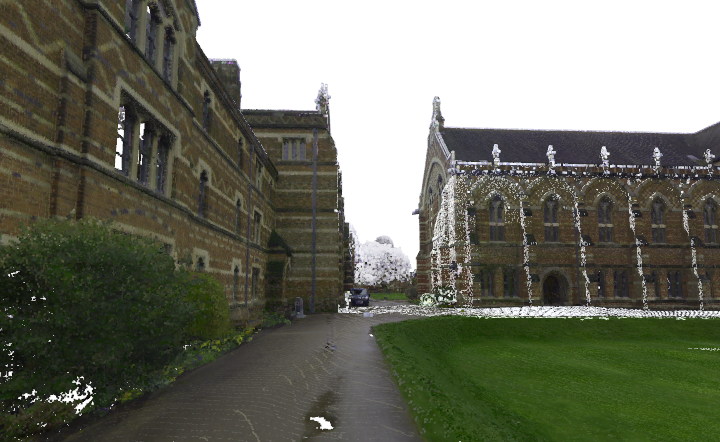}
\end{subfigure}%
\vspace{1ex}
\begin{subfigure}[b]{1.0\linewidth}
    \centering
    \captionsetup{justification=centering}
    \includegraphics[width=0.99\textwidth]{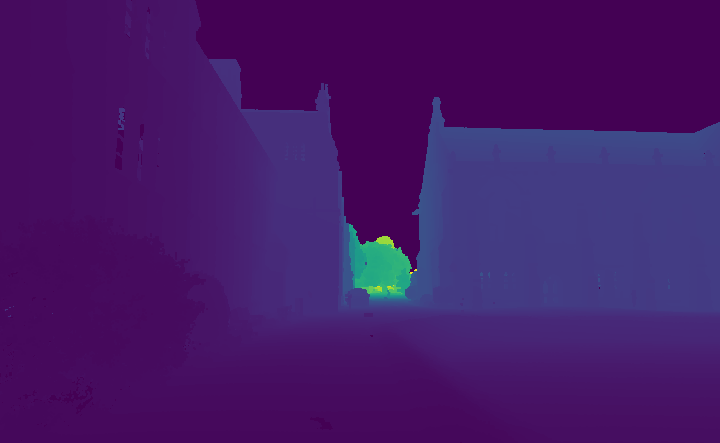}
\end{subfigure}%

\caption{\textit{Top}: Image rendered from TLS map in Keble College. \textit{Bottom}: Depth image corresponding to the rendered image.}
\label{fig:rendered}
\end{figure}

\subsubsection{Processed - COLMAP Outputs:}
A solution to SfM is required as input to both MVS methods and radiance field methods (NeRF and 3D Gaussian Splatting). To facilitate researchers, we ran the state-of-the-art SfM method COLMAP~\citep{schoenberger2016colmap} for each sequence and provide its outputs. Specifically, COLMAP provides camera information in \texttt{cameras.bin}, image information in \texttt{images.bin}, 3D feature points in \texttt{points3D.bin} and the database information in \texttt{database.db}. 

Running SfM for all images captured at 20 Hz results in a large amount of output data and computation time. This is unnecessary because consecutive images are very close to each other and thus redundant. To keep the number of images manageable yet providing enough viewpoints for visual reconstruction, we selected images that are synchronised to the SLAM pose graph point clouds and spaced \SI{1}{\meter} apart. We then ran COLMAP on this set of images. At walking speed, this results in a frequency of about 1 Hz. For each sequence, the total number of images was less than 2000 (for each of the three cameras). Using images aligned with corresponding LiDAR depth is also useful for methods that fuse LiDAR and vision, for example, colourising point clouds and depth-aided radiance fields such as Urban Radiance Field~\citep{rematas2022urban}. These aligned depth images are also provided in addition to the COLMAP outputs.

For compatibility with Nerfstudio~\citep{tancik2023nerfstudio} (a popular open-sourced code base for state-of-the-art radiance field methods), we also convert the outputs from COLMAP into a \texttt{transforms.json} file. Specifically, \texttt{transforms.json} includes camera parameters (camera models, focal length, principle point, image size, distortion parameters), image file path and the corresponding $\mathit{SE}(3)$ pose estimate as a $4\times4$ transformation matrix. 

Correcting the metric scale of vision-based 3D reconstructions produced by MVS and radiance field methods is necessary to enable comparison to the metric ground truth. To estimate the scale, we used Umeyama's method\footnote{We used implementation from \url{https://github.com/MichaelGrupp/evo}} to estimate a $\mathit{Sim}(3)$ transformation between the LiDAR trajectory and a COLMAP trajectory, and the results are saved in \texttt{evo\_align\_results.json}. We provide tools to compute the scale parameter and to rescale the MVS reconstruction and radiance field reconstruction to metric size.

\subsubsection{Ground Truth - Reconstruction:}
\label{sec:gt_recon}
We provide the registered individual TLS scans from Leica RTC360 for each site as the ground truth reconstruction. Each scan is saved as \texttt{<site-00x>.e57} under each site folder, and contains not only the point clouds but also the sensor origin, which is important in reconstruction methods such as occupancy mapping~\citep{hornung13octomap}. Moreover, we also provide the complete colourised TLS map at \SI{1}{\centi\meter} resolution for each site (\figref{fig:bodleian}) by merging the individual RTC360 scans. The TLS scan registration reports from the Leica software are provided as \texttt{FinalizeReport.pdf} in each ground truth site folder, which specifies the number of scans, overlap, residual in the merging process, etc. Regarding the density, we tried to distribute the scans uniformly on each site.

\subsubsection{Ground Truth - Localisation:}
\label{sec:loc_gt}
The ground truth trajectory is computed by ICP registering each undistorted LiDAR point cloud (as described in \secref{sec:slam_outputs}) to the merged TLS map described in \secref{sec:gt_recon}. To obtain the ground truth pose for each LiDAR scan, we use an offline version of VILENS~\citep{wisth2023vilens} with Iterative Closest Point (ICP)~\citep{besl1992method} at its core. Running offline, we can allocate enough time to register the LiDAR to the colourised TLS map. The trajectory estimated through the described process is synchronised with cameras by the procedure described in \secref{sec:hardware_perception}. This is in the same manner as for Newer College~\citep{ramezani2020newer} and Hilti-2022~\citep{zhang2022hilti}. The accuracy of the ground truth trajectory is approximately 1-2~\SI{}{\centi\meter}. We validated the ground truth trajectory by projecting the individual LiDAR scans into a map and comparing them to the TLS map. The trajectory is provided as \texttt{gt-tum.txt} in TUM~\citep{sturm2012slambenchmark} format, with each line encoding timestamp, position (x,y,z) and quaternion (x,y,z,w).

\subsubsection{Depth images from the TLS map:}
\label{sec:depth}
Additionally, we include ground truth depth images rendered from the TLS map using the ground truth sensor trajectories. In \figref{fig:rendered}, we show an example of an image rendered from the Keble College site with its corresponding depth image. These images could be used for evaluating monocular depth estimation or novel view rendering.

\begin{figure*}[t]
\centering
\begin{subfigure}[b]{0.25\linewidth}
    \centering
    \captionsetup{justification=centering}
    \includegraphics[width=0.99\textwidth]{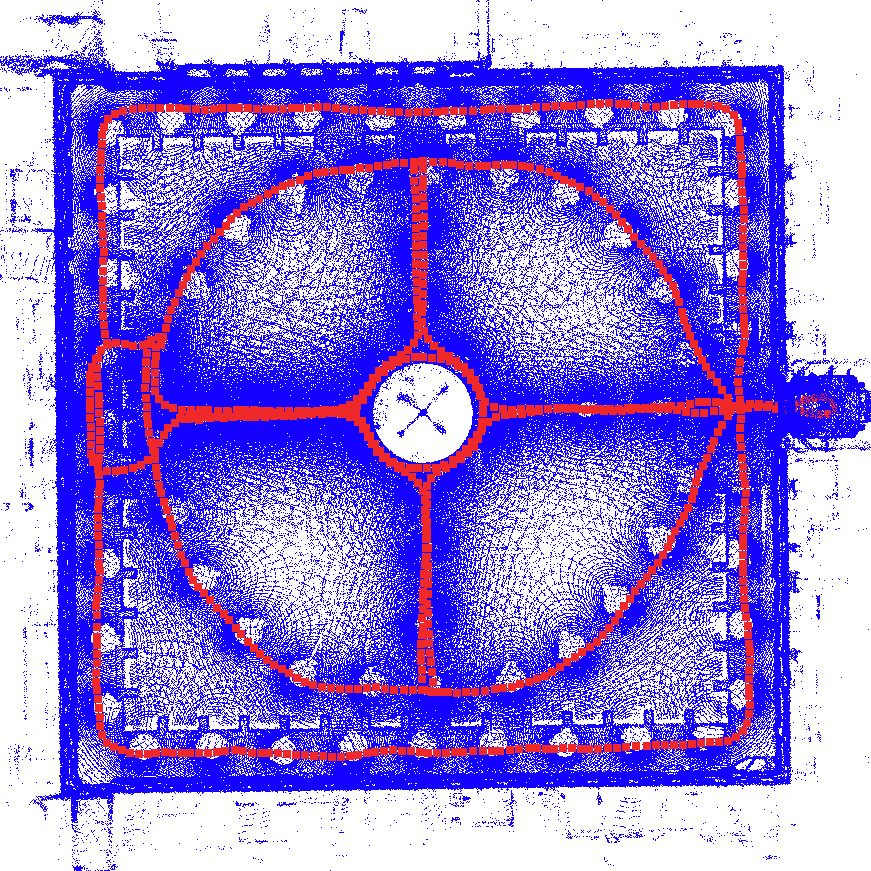}
    \caption{Christ Church College}
\end{subfigure}%
\begin{subfigure}[b]{0.25\linewidth}
    \centering
    \captionsetup{justification=centering}
    \includegraphics[width=0.99\textwidth]{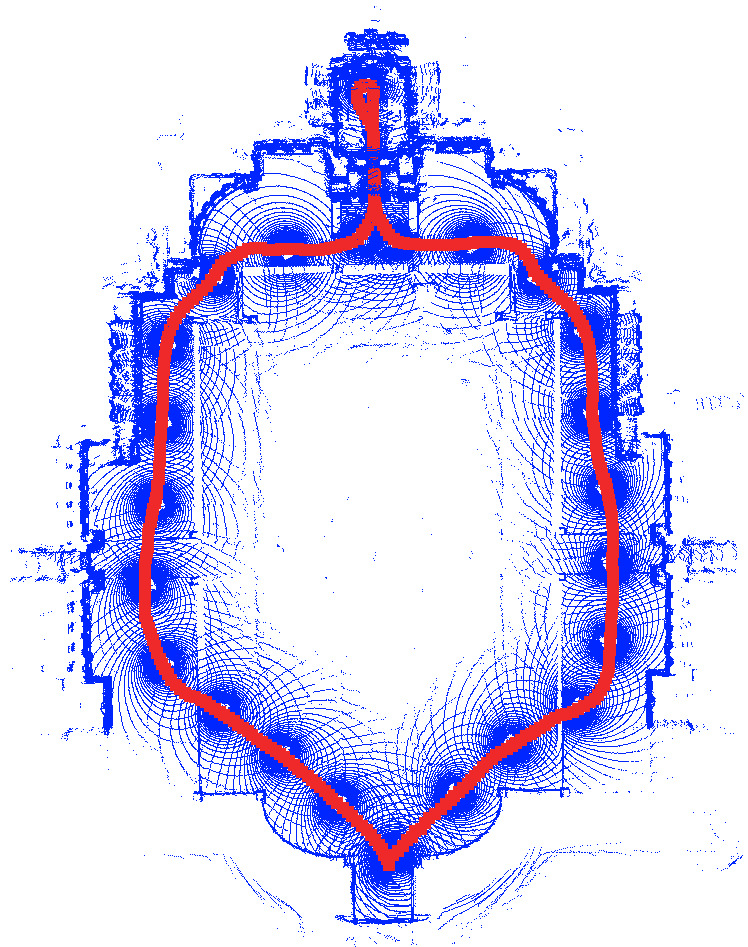}
    \caption{Blenheim Palace}
\end{subfigure}%
\begin{subfigure}[b]{0.25\linewidth}
    \centering
    \captionsetup{justification=centering}
    \includegraphics[width=0.99\textwidth]{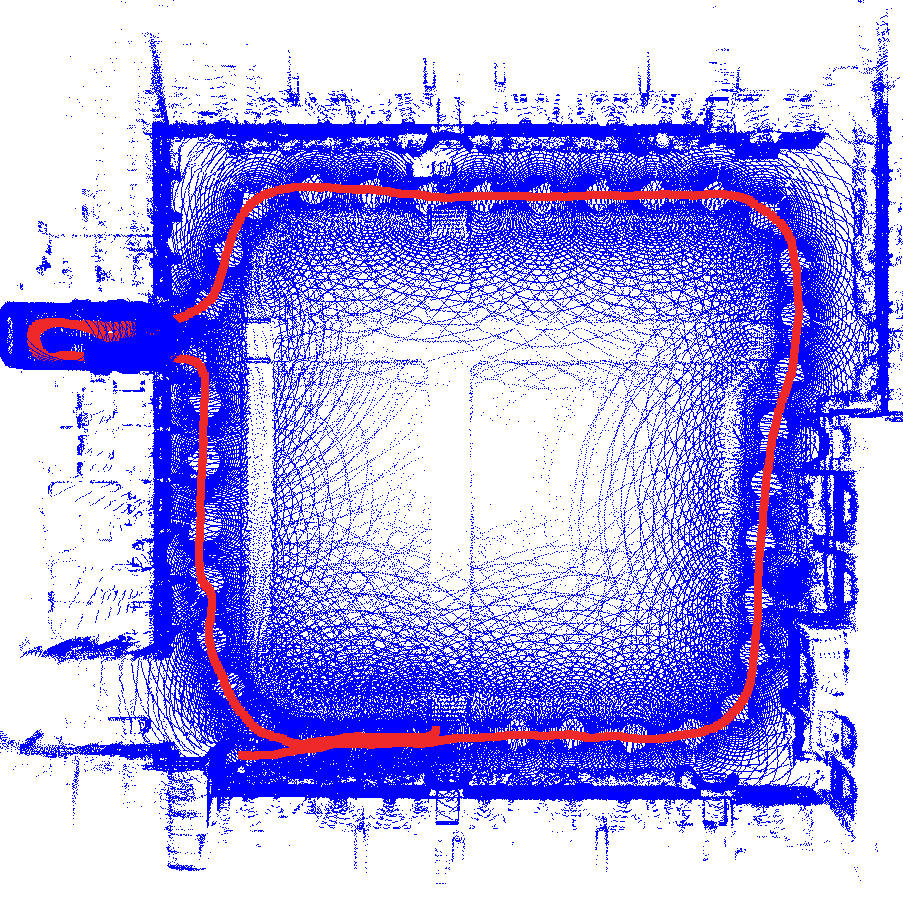}
    \caption{Keble College}
\end{subfigure}%
\begin{subfigure}[b]{0.25\linewidth}
    \centering
    \captionsetup{justification=centering}
    \includegraphics[width=0.99\textwidth]{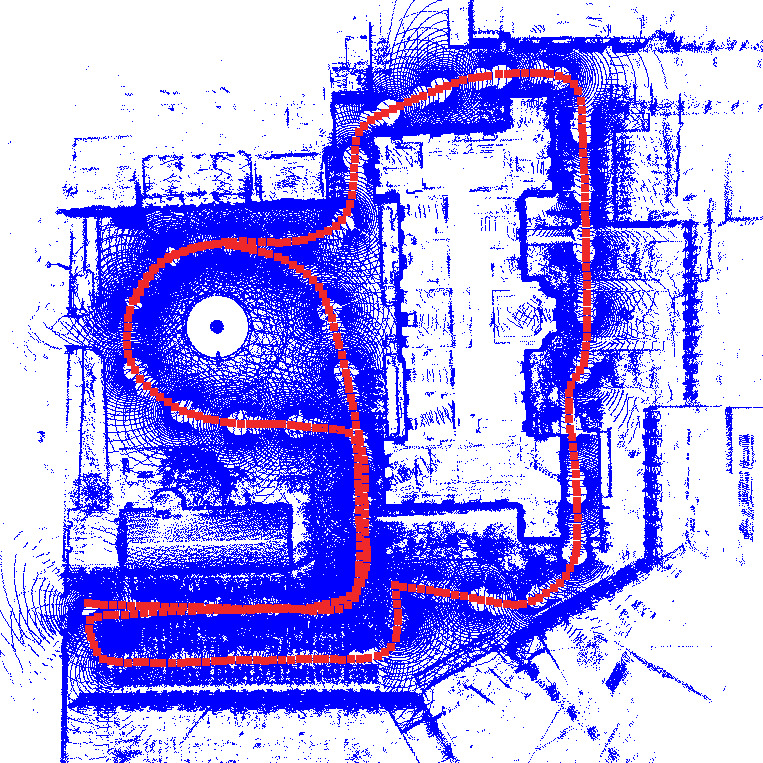}
    \caption{Radcliffe Observatory Quarter}
\end{subfigure}%
\caption{Examples of SLAM trajectories (in red) and LiDAR point cloud maps (in blue) for four sequences from the dataset.}
\label{fig:mosaic_seq}
\end{figure*}

\subsection{Sequence Description}
\label{sec:sequences}
The dataset was recorded in six historic sites in Oxford, UK:
\begin{itemize}
\item Bodleian Library ($\sim$\SI{37000}{\meter^2})
\item Blenheim Palace ($\sim$\SI{14000}{\meter^2})
\item Christ Church College ($\sim$\SI{26000}{\meter^2})
\item Keble College ($\sim$\SI{18000}{\meter^2})
\item Radcliffe Observatory Quarter (ROQ) ($\sim$\SI{12000}{\meter^2})
\item New College ($\sim$\SI{18000}{\meter^2})
\end{itemize}

Each sequence was collected by walking with the Frontier payload device mounted in a backpack as shown in \figref{fig:frontier_mounted}. 

\subsubsection{Bodleian Library:}
This site consists of the area around the Bodleian Library, which includes Radcliffe Square, where the Radcliffe Camera and Oxford's University Church are located. It also reconstructs the outside of the Sheldonian Theatre and some of Broad Street. This part of the dataset contains the most iconic landmarks of the historic centre of Oxford (\figref{fig:bodleian}).

For this site, we provide two outdoor trajectories of walking through streets and squares around the described area. The recordings contain many details of the predominant medieval and Gothic buildings.

\subsubsection{Blenheim Palace:}
This is one of England's largest houses and is notable as Sir Winston Churchill's ancestral home. Five trajectories were captured in the palace's main square, the principal hall, and rooms in the west wing, including the library. The trajectories have outdoor and indoor parts, including different-sized rooms and corridors. In \figref{fig:mosaic_seq} (b), we show an example of a sequence in the palace's main square.

\subsubsection{Christ Church College:}
Founded in 1546, Christ Church is a constituent college of Oxford and one of the city's best-recognised locations. It contains Tom Quad, the largest square in Oxford, the college dining hall as well as Christ Church Cathedral and its cloister.

The sequences recorded in this site include outdoor areas with pavements and lawns as well as indoor parts with different lighting conditions and stairs accessing different levels, including the dining hall. One sequence is shown in \figref{fig:mosaic_seq} (a), which includes a complete loop of the perimeter of Tom Quad, which is challenging due to the limited range of the LiDAR sensor and repeating architecture.

\subsubsection{Keble College:}
Keble is another constituent college of the University of Oxford. It comprises neo-Gothic-style buildings, including a hall and a church. Keble's buildings are distinctive because they are constructed of alternating red and white coloured bricks --- which provides an interesting challenge to visual reconstruction. This differentiates it from the other sites which are mostly built from limestone.

The Keble sequences were recorded outdoors in the college's squares (\figref{fig:mosaic_seq} (c)), which include lawns and trees, as well as some interior and exterior parts.

\subsubsection{Radcliffe Observatory Quarter:}
The ROQ site consists of the Faculty of Philosophy, the Mathematical Institute, and St Luke's Chapel. This area is near the Oxford Robotics Institute, where the authors are affiliated. The two sequences recorded here contain squares with pavement, lawns, trees, narrow spaces between some buildings, and a fountain containing fine 3D details. In \figref{fig:mosaic_seq} (d), we show an example of a sequence through this site.

\subsubsection{New College:}
New College is another constituent college of the University of Oxford and is located in the city's historic centre. It contains squares, a hall, a church and a cloister. We recorded four sequences in New College, including an oval lawn area at the centre of the main quad surrounded by medieval buildings. The sequences combine outdoor and indoor parts with abrupt changes in light conditions. Most of the 03 sequence was walking through the park, containing a lawn and many trees. Some parts were fully covered by tree canopies. This site corresponds to the earlier New College Dataset~\citep{smith2009new,ramezani2020newer}.


\section{Benchmarks and Results}
In this section, we describe three benchmarks we have created to demonstrate our dataset.
The benchmarks compare state-of-the-art methods for localisation (\secref{sec:localisation}), 3D reconstruction (\secref{sec:3d_recon}), and novel view synthesis~(\secref{sec:novel_view}).

\begin{table*}[t]
\footnotesize
    \centering
    \caption{Results in RMS (m) of the ATE  using the provided ground truth. We mark the best results in blue using different tints. SC-LIO-SAM fails on some sequences. COLMAP gives incomplete results on some sequences.}
    \begin{tabular}{c c c||c c c c ||c||c c }
    \toprule
         \textbf{Site} & \textbf{Sec} & \textbf{Len} & \textbf{VILENS-SLAM} & \textbf{Fast-LIO-SLAM} & \textbf{SC-LIO-SAM} & \textbf{ImMesh} & \textbf{Fast-LIVO2} & \textbf{HBA} & \textbf{COLMAP} \\
    \midrule
         Keble College & 02 & 290 & \cellcolor{top2}0.06 & 0.25 & 1.26 & \cellcolor{top3}0.08 & 0.95 & 0.11 & \cellcolor{top1}0.05 \\
                       & 03 & 280 & 0.14 & \cellcolor{top3}0.11 & 4.02 & 0.14 & \cellcolor{top2}0.06 & 0.12 & \cellcolor{top1}0.05  \\
                       & 04 & 780 & 0.16 & 0.49 & \xmark & 3.67 & \cellcolor{top2}0.09 & \cellcolor{top3}0.12 & \cellcolor{top1}0.07 \\
                       & 05 & 710 & \cellcolor{top2}0.11 & 0.29 & \xmark & \cellcolor{top3}0.13 & \cellcolor{top2}0.11 & \cellcolor{top3}0.13 & \cellcolor{top1}0.09 \\
    \midrule
         Radcliffe Obs.  & 01 & 400 & \cellcolor{top3}0.06 & 0.17 & 0.23 & 0.20  & \cellcolor{top1}0.04 & \cellcolor{top2}0.05 & 0.07 \\
         Quarter         & 02 & 390 & \cellcolor{top3}0.09 & 0.24 & 0.14 & 0.27  & \cellcolor{top1}0.07 & \cellcolor{top2}0.08 & \cellcolor{top2}0.08 \\
    \midrule
         Blenheim Palace & 01 & 490 & 0.47 & \cellcolor{top3}0.18  & 6.74 & 0.27  & \cellcolor{top2}0.14 & 0.21 & \cellcolor{top1}0.08 \\
                         & 02 & 390 & \cellcolor{top3}0.16 & 0.12 & 4.41 & 0.36  & 0.22 & \cellcolor{top2}0.08 & \cellcolor{top1}0.05 \\
                         & 05 & 390 & 1.05 & 0.28 & \xmark & \cellcolor{top2}0.22  & \cellcolor{top3}0.26 & \cellcolor{top1}0.14 & \cellcolor{top3}0.26 \\
    \midrule
         Christ Church & 01 & 920 & \cellcolor{top1}0.06 & 0.72 & \xmark & 0.19  & 0.54 & \cellcolor{top3}0.07 & \cellcolor{top1}0.06 \\
         College       & 02 & 640 & \cellcolor{top3}0.17 & 0.49 & \xmark & 1.70  & 0.63 & \cellcolor{top1}0.12 & \cellcolor{top2}0.15 \\
                       & 03 & 340 & \cellcolor{top1}0.03 & 0.23 & 0.14 & 0.16  & \xmark & \cellcolor{top2}0.05 & \cellcolor{top3}0.07 \\
                       & 05 & 820 & \cellcolor{top3}0.17 & 0.30 & \xmark & 0.21  & \cellcolor{top2}0.15 & \cellcolor{top1}0.12 & \xmark \\
    \midrule
        Bodleian Library
                          & 02 & 690 & 1.11 & \cellcolor{top1}{0.25} & 1.71 & \cellcolor{top3}{0.39}  & 0.46 & 0.89 & \cellcolor{top2}{0.27} \\
    \bottomrule
    \end{tabular}
    \label{tab:ate_results}
\end{table*}

\subsection{Localisation Benchmark}
\label{sec:localisation}

In the localisation benchmark, we evaluate the trajectories estimated for each sequence using state-of-the-art LiDAR Inertial SLAM and SfM approaches:
\begin{itemize}
    \item VILENS-SLAM: VILENS~\citep{wisth2023vilens} with pose graph optimisation~\citep{ramezani2020slam} (online).
    \item Fast-LIO-SLAM~\citep{kim2022sc}: Fast-LIO2~\citep{xu2022fast} with pose graph optimisation and Scan Context loop closures~\citep{kim2018scan} (online).
    \item SC-LIO-SAM~\citep{kim2022sc}: LIO-SAM~\citep{shan2020liosam} with pose graph optimisation and Scan Context loop closures~\citep{kim2018scan} (online).
    \item ImMesh~\citep{lin2023immesh}: LiDAR meshing with Fast-LIO2 odometry (online).
    \item Fast-LIVO2~\citep{zheng2024fast}: LiDAR visual inertial odometry (online).
    \item HBA~\citep{liu2023hba}: LiDAR bundle adjustment using as input the VILENS-SLAM result (offline).
    \item COLMAP~\citep{schoenberger2016colmap}: Structure-from-Motion using only images (offline). For image matching, we used a sequential matcher with loop closure detection.
\end{itemize}

We evaluate sequences on the trajectories which were completely within the ground truth TLS map. We exclude Keble College 01, Blenheim Palace 03 and 04, Christ Church College 04 and 06, and Bodleian Library 01 from this analysis.

\begin{figure}[ht]
\centering \includegraphics[width=1.0\columnwidth]{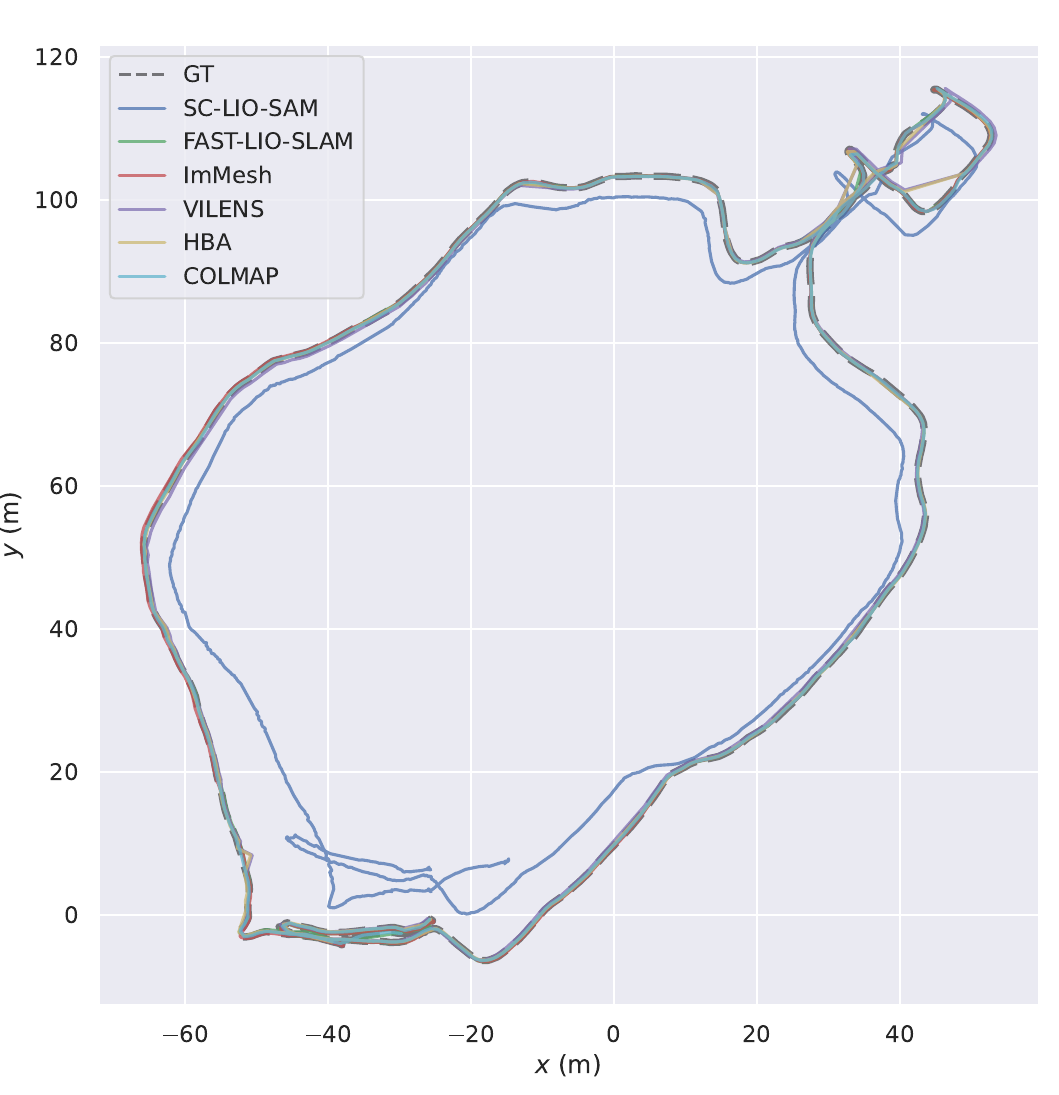}
	\caption{A top-down view showing a representative performance of the different systems for Sequence 01 at Blenheim Palace. The sequence starts and ends in the lower left. The environment where this sequence was collected can be seen in \figref{fig:mosaic_seq} (b).}
	\label{fig:trajactory_xy}
\end{figure}

\subsubsection{Evaluation Metrics:}

In the dataset tools, we provide a Python script to compute Absolute Trajectory Error (ATE) and Relative Pose Error (RPE) metrics. We used ATE to compare the poses estimated by the SLAM and SfM methods as they should be globally consistent (\tabref{tab:ate_results}). Even though Fast-LIVO2 is an odometry system, its drift rate is low. Thus, we consider it globally consistent and also evaluated it with ATE.

To transform the trajectories estimated by the methods to the ground truth coordinate frame (\secref{sec:loc_gt}), we use the $\mathit{SE}(3)$ Umeyama alignment. For comparison between odometry systems, we recommend using RPE to measure local performance.

\subsubsection{Experimental Results:}
\label{sec:eval_loc}

A comparison between the listed methods using the provided ground truth (\secref{sec:loc_gt}) is presented in \tabref{tab:ate_results} for each sequence. We mark the best results in dark blue, while the second and third are highlighted with a lighter tint. The offline methods on the right side (HBA and COLMAP) can take advantage of all available data and provide the most accurate results. While one might expect that HBA should provide better accuracy (as a LiDAR bundle adjustment approach), we achieved the best performance with COLMAP. We believe this is due to the overlapping multicamera configuration, which provides abundant view constraints and allows features to be seen from different perspectives with proper light conditions in general. Furthermore, the prior map we provide to HBA is imperfect which means that the undistorted point clouds can be imperfect, which would then lead in turn to inaccuracies in HBA. However, note also that COLMAP was unable to solve some sequences (with poor lighting) at all and instead produced multiple disconnected sub-models. This is usually due to insufficient visual features being matched in the area where two sub-models ought to connect, which can be due to there being insufficient visual overlap or there being challenging lighting conditions (e.g., The dining hall in Christ Church College is relatively dark). In comparison, the LiDAR SLAM systems are invariant to lighting conditions and the present visual features. 

The second most accurate method is HBA, which further refines VILENS-SLAM's trajectory estimation with LiDAR bundle adjustment in post-processing.
Of the online methods, VILENS-SLAM and Fast-LIVO2 performed best. Fast-LIVO2 performs better in short-loop sequences (Keble College and Radcliff Observatory Quarter), as the map created is also seen, and its drift is low. In contrast, VILENS-SLAM performs better in large loops as it performs loop closures. ImMesh and Fast-LIO-SLAM use Fast-LIO2~\citep{xu2022fast} as their core odometry module, which achieved accurate trajectory estimation. SC-Fast-LIO performs loop closure correction using Scan Context~\citep{kim2018scan}, while ImMesh can recover from drifts while generating the mesh map. SC-LIO-SAM produces satisfactory results on some sequences using Scan Context~\citep{kim2018scan} as an appearance-based place recognition module. However, it adds incorrect loop closures in sequences with large loops and repeated building patterns, such as Christ Church College and Blenheim Palace.
We note that these methods could potentially perform better with further parameter tuning.

In \figref{fig:trajactory_xy}, we show a representative example of the performance of the evaluated methods using Sequence 01 of Blenheim Palace. All of the methods produce an accurate trajectory except for SC-LIO-SAM, which incorporates an incorrect loop closure when closing the large loop. 


\begin{table*}[t]
    \footnotesize
    \caption{\small{Quantitative evaluation of the 3D reconstructions from VILENS-SLAM, OpenMVS and Nerfacto. We indicate the best results with a dark blue background.}}
    \centering
    \begin{tabular}{c c l|c c|c c c|c c c}
    \toprule
    \multirow{2}{*}{Site}&\multirow{2}{*}{SEC.}&\multirow{2}{*}{Method} &\multirow{2}{*}{Accuracy$\downarrow$} & \multirow{2}{*}{Completeness$\downarrow$} &\multicolumn{3}{c|}{5cm}&\multicolumn{3}{c}{10cm}\\
    &&&&&Precision& Recall & F-score&Precision& Recall & F-score\\
    \hline
    \addlinespace    
    \multirow{3}{0.06\linewidth}{Blenheim Palace}&\multirow{3}{*}{05}
    &VILENS-SLAM	&\cellcolor{top1}0.070	&\cellcolor{top1}0.506&\cellcolor{top1}0.670&	\cellcolor{top1}0.392&	\cellcolor{top1}0.495	&\cellcolor{top1}0.867	&\cellcolor{top1}0.661	&\cellcolor{top1}0.750 \\
    &&OpenMVS& \cellcolor{top2}0.126	& 1.045 &\cellcolor{top2}0.451&	\cellcolor{top2}0.251&	\cellcolor{top2}0.323&\cellcolor{top2}	0.574	&\cellcolor{top2}0.381&	\cellcolor{top2}0.458\\
    &&Nerfacto & 0.302	&\cellcolor{top2}0.676&  0.232	& 0.094	& 0.134&	 0.388& 	0.257& 	0.309\\
    \hline
    \addlinespace
    \multirow{3}{0.06\linewidth}{Christ Church College}&\multirow{3}{*}{02}
    &VILENS-SLAM	&\cellcolor{top2}0.082	&\cellcolor{top1}3.296& \cellcolor{top2}0.540&	\cellcolor{top1}0.250&	\cellcolor{top1}0.342	&\cellcolor{top2}0.794	&\cellcolor{top1}0.408	&\cellcolor{top1}0.539\\
    &&OpenMVS &\cellcolor{top1} 0.046&	 5.381 &\cellcolor{top1}0.771&	\cellcolor{top2}0.201	&\cellcolor{top2}0.319&	\cellcolor{top1}0.886&	\cellcolor{top2}0.266&	\cellcolor{top2}0.410\\
    &&Nerfacto & 0.219	&\cellcolor{top2}4.435&  0.328	& 0.157	& 0.212& 	0.532	& 0.254&	 0.343\\
    \hline
    \addlinespace
    \multirow{3}{0.06\linewidth}{Keble College}&\multirow{3}{*}{04}
    &VILENS-SLAM	&\cellcolor{top2} 0.067	&\cellcolor{top2}0.342 &\cellcolor{top2}0.527&	\cellcolor{top2}0.527&	\cellcolor{top2}0.527&	\cellcolor{top2}0.816&	\cellcolor{top1}0.779&	\cellcolor{top2}0.797\\
    &&OpenMVS &\cellcolor{top1} 0.050&	 0.409 &\cellcolor{top1}0.766	&\cellcolor{top1}0.606	&\cellcolor{top1}0.677&\cellcolor{top1}	0.918&	\cellcolor{top2}0.718&	\cellcolor{top1}0.806\\
    &&Nerfacto & 0.137&	\cellcolor{top1}0.150&  0.418	& 0.484	& 0.449	& 0.654	& 0.709	& 0.680\\
    \hline
    \addlinespace

    \multirow{3}{0.06\linewidth}{Radcliffe Obs. Quarter}&\multirow{3}{*}{01}
    &VILENS-SLAM	&\cellcolor{top1}0.047	&\cellcolor{top1}0.233 &\cellcolor{top2}0.708	&\cellcolor{top1}0.536&\cellcolor{top1}	0.610	&\cellcolor{top1}0.909	&\cellcolor{top1}0.806	&\cellcolor{top1}0.854\\
    &&OpenMVS &\cellcolor{top2}0.048&	 0.622 &\cellcolor{top1}0.745	&\cellcolor{top2}0.470&	\cellcolor{top2}0.577	&\cellcolor{top2}0.902&	\cellcolor{top2}0.618	&\cellcolor{top2}0.734\\
    &&Nerfacto &  0.197&\cellcolor{top2}	0.398&  0.415	& 0.395& 	0.405	& 0.587	& 0.598	& 0.592\\
    \bottomrule
    \addlinespace
    \end{tabular}
    \label{tab:reconstruction_eval}
\end{table*}

\begin{figure*}[h]
    \centering 
    \includegraphics[width=2\columnwidth]{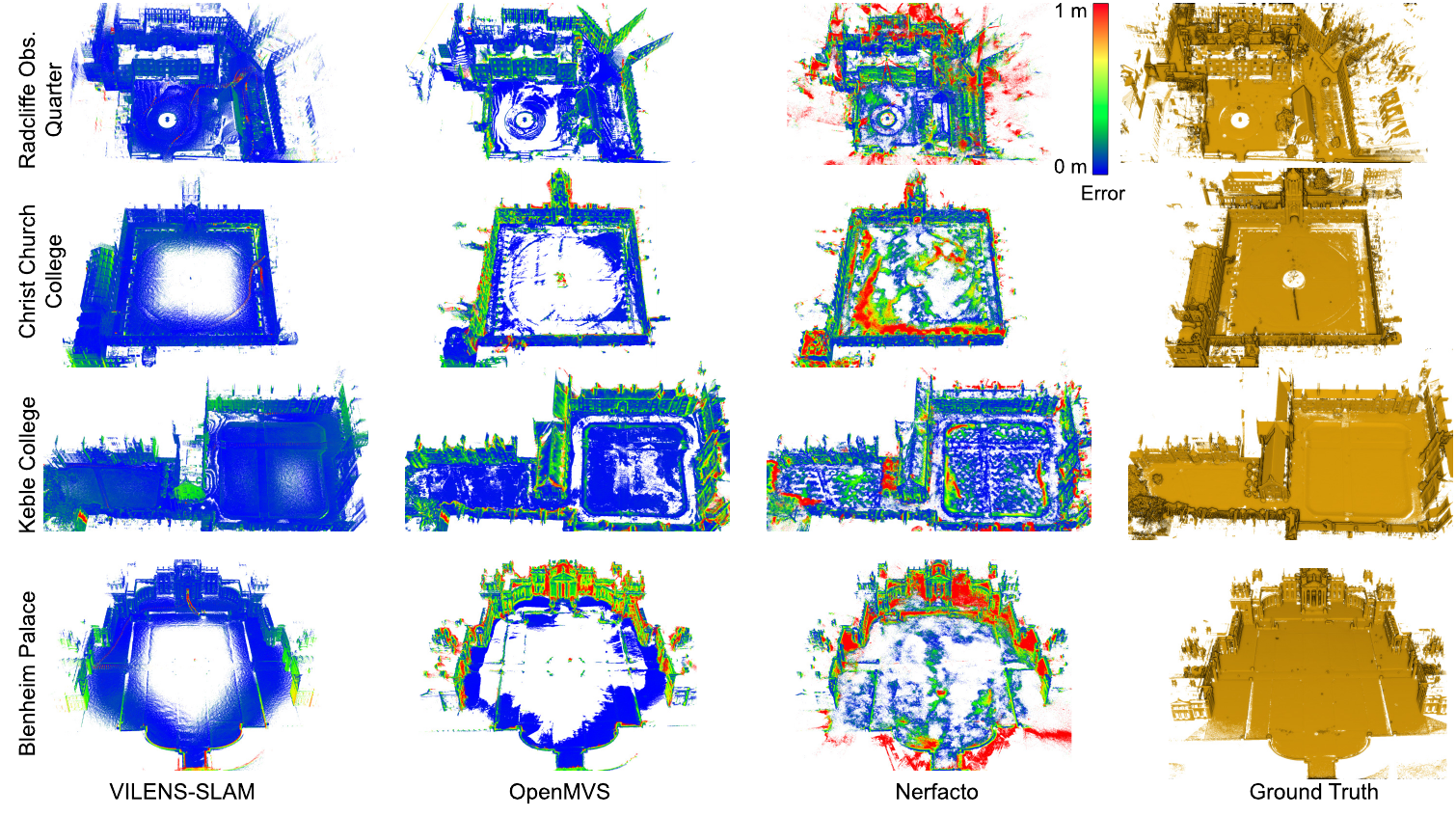}
    \caption{Comparison between the reconstructions achieved by the different methods. The reconstructions in the first three columns are coloured by point-to-point distance to the ground truth model. }
    \label{fig:recon_compare}
\end{figure*}

\subsection{3D Reconstruction Benchmark}
\label{sec:3d_recon}
The reconstruction benchmark evaluates outputs from the systems that use vision or LiDAR. Specifically, we evaluate the following systems:
\begin{itemize}
    \item VILENS-SLAM: Merged LiDAR point clouds using poses obtained using the LiDAR SLAM system described in \secref{sec:localisation} (online).
    \item OpenMVS\footnote{\url{https://cdcseacave.github.io/openMVS}}: an MVS system which uses input from COLMAP~\citep{schoenberger2016colmap} (offline).
    \item Nerfacto: The default and recommended method from Nerfstudio\footnote{We used version 1.1.4}~\citep{tancik2023nerfstudio} that combines features from MipNeRF-360~\citep{barron2022mipnerf360}, Instant-NGP~\citep{muller2022instant} and others. It uses input from COLMAP, as with OpenMVS (offline).
\end{itemize}

The outputs from each system are all in the form of 3D point clouds. For Nerfacto, the point cloud is generated from the trained model by calculating the expected depth and colour for the training rays, and projecting the depth points into 3D. 

We selected example trajectories that are completely within the ground truth reconstruction from Blenheim Palace, Christ Church College, Keble College and Radcliffe Observatory Quarter.

\subsubsection{Evaluation Metrics:}
We use the F-score as the primary metric for reconstruction. The F-score is calculated as the harmonic mean of precision and recall, thus it considers both aspects of the reconstruction: accuracy and completeness. To calculate precision and recall, we consider a point to be a true positive (TP) if the distance from it to the closest ground truth point is within a certain threshold. We report results using \SI{5}{\centi\meter} and \SI{10}{\centi\meter} thresholds. False positives (FP) are reconstructions that are further from the ground truth and thus inaccurate. False negatives (FN) are regions in the ground truth that have no neighbouring points in the reconstruction, and are thus incomplete. Specifically, precision and recall are defined by

\begin{equation}
    \begin{aligned}
        \text{Precision} = \frac{TP}{TP + FP}
    \end{aligned}
\end{equation}

\begin{equation}
    \begin{aligned}
        \text{Recall} = \frac{TP}{TP + FN}
    \end{aligned}
\end{equation}

The F-score is then calculated as 

\begin{equation}
    \begin{aligned}
        F_1 = 2 \cdot \frac{\text{Precision} \cdot \text{Recall}}{\text{Precision} + \text{Recall}}
    \end{aligned}
\end{equation}
We also report the point-to-point distances to measure accuracy and completeness following the conventions of the DTU dataset~\citep{aanaes2016large_dtu}. Specifically, accuracy
is measured as the point-to-point distance from the reconstruction to the ground truth map and measures
the reconstruction quality (the lower, the better). Completeness is the distance from
the point-wise reference map to the reconstruction and indicates how
much of the ground truth surface has been captured by the reconstruction (the lower, the better). However, these metrics are more sensitive to outliers and non-overlapping regions, which may skew the results in practice. Thus, we do not use them as the primary metrics.

\subsubsection{Reconstruction Filtering:}
\label{sec:recon_pre}
In practice, the reconstruction and ground truth reference models will contain regions which were not mutually scanned. If unaccounted for, this would lead to erroneous false positives and false negatives. In turn, this would result in precision and recall measures which do not reflect the true quality of the reconstruction. For fairer comparison, we filter out points in the reconstruction that fall outside the reconstructed ground truth region, i.e. the regions not reconstructed in the ground truth model. In particular, for Nerfacto the sky must be specifically removed because, as a dense representation, it attempts to reconstruct it using available depth cues. We filter these sky point clouds to ensure the evaluation focuses on the reconstruction of the physical environment itself.

\begin{table*}[t]
    \small
    \caption{\small{Quantitative evaluation of Novel View Synthesis. The best results are coloured in blue using different tints. The test images are selected from the  input trajectory (In-Sequence) as well as a separate trajectory with viewpoints far from the input trajectory (Out-of-Sequence).}}
    \centering
    \begin{tabular}{l l | c c c c| c c c c}
    \toprule
    \multirow{2}{4em}{Sequence}&\multirow{2}{4em}{Method}&\multicolumn{3}{c}{In-Sequence}&&&\multicolumn{3}{c}{Out-of-Sequence}\\
    &&PSNR$\uparrow$ &SSIM$\uparrow$&LPIPS$\downarrow$&&& PSNR$\uparrow$&SSIM$\uparrow$&LPIPS$\downarrow$\\

    \midrule

    \multirow{4}{0.06\linewidth}{Observatory Quarter}
    &Nerfacto &\cellcolor{top2}23.40&\cellcolor{top2}0.807&\cellcolor{top2}0.336&&&\cellcolor{top1}21.25&\cellcolor{top2}0.786&\cellcolor{top2}0.370\\
    &Nerfacto-big&20.66&\cellcolor{top2}0.807&\cellcolor{top1}0.292&&&19.38&\cellcolor{top1}0.787&\cellcolor{top1}0.317\\
    &Splatfacto& 22.76&0.791&0.373&&& 19.47&0.736&0.445\\
    &Splatfacto-big&\cellcolor{top1}23.54&\cellcolor{top1}0.811& 0.347&&&\cellcolor{top2}20.26& 0.761& 0.413\\
    
    \midrule
    
    \multirow{4}{0.06\linewidth}{Blenheim Palace}
    &Nerfacto&  18.42&0.716&\cellcolor{top2}0.506&&&\cellcolor{top1}17.09&\cellcolor{top2}0.682&\cellcolor{top2}0.537\\
    &Nerfacto-big&17.93& 0.724&\cellcolor{top1}0.445&&&\cellcolor{top1}17.09&\cellcolor{top1}0.695&\cellcolor{top1}0.493\\
    &Splatfacto&\cellcolor{top2}19.34&\cellcolor{top2}0.726&0.589&&&16.02&0.668&0.659\\
    &Splatfacto-big&\cellcolor{top1}19.77&\cellcolor{top1}0.733& 0.576&&& 16.20& 0.671& 0.643\\

    \midrule

    \multirow{4}{0.06\linewidth}{Keble College}
    &Nerfacto&\cellcolor{top2}21.10&\cellcolor{top2}0.731&\cellcolor{top2}0.397&&&\cellcolor{top2}20.29&\cellcolor{top1}0.748&\cellcolor{top1}0.368\\
    &Nerfacto-big&19.71&\cellcolor{top1}0.749&\cellcolor{top1}0.326&&&18.15&\cellcolor{top2}0.736&\cellcolor{top2}0.381\\
    &Splatfacto& 20.47&0.651&0.514&&& 19.92&0.658&0.500\\
    &Splatfacto-big&\cellcolor{top1}21.36& 0.688& 0.478&&&\cellcolor{top1}20.86& 0.707& 0.434\\

    \bottomrule
    \addlinespace
    \end{tabular}
    \label{tab:nerf_eval}
\end{table*}

\subsubsection{Experimental Results:}

The quantitative evaluations of the reconstructions are presented in \tabref{tab:reconstruction_eval}, and the qualitative results are shown in \figref{fig:recon_compare}. VILENS-SLAM reconstruction achieves the best F-score in most experiments except Keble-04. This is a reasonable result given LiDAR's accurate depth measurements. The remaining inaccuracy mostly comes from trajectory errors in long sequences and the presence of dynamic objects. The reconstruction completeness is limited by the short sensor range in Christ Church College and Blenheim where the central region of the large squares is not reconstructed.

Reconstructions from OpenMVS are accurate in regions with abundant view constraints and distinct texture, but it is not able to reconstruct surfaces with uniform texture such as the ground in Blenheim Palace and the lawn in Christ Church College. The error distribution in the MVS cloud is not uniform and tends to appear at surface boundaries where occlusion is an issue.

Although both OpenMVS and Nerfacto are purely vision-based reconstruction methods, Nerfacto point clouds are generally less precise. This is because MVS filters uncertain points (by checking photo-consistency), but the NeRF approach instead optimises a continuous radiance field without an explicit notion of uncertainty. For regions with insufficient view constraints and uniform texture, Nerfacto estimates incorrect depth values which leads to uneven ground reconstructions. In comparison, OpenMVS filters some of the reconstruction there, which leads to better precision and accuracy. 

The reconstruction quality is determined not only by the reconstruction method but also by the accuracy of the input trajectory. Both precision and recall can be affected by an imperfect trajectory estimation. Clouds produced by VILENS-SLAM contain surfaces with high error that are the result of incorrectly registered LiDAR scans. Meanwhile, for Christ Church College, both the OpenMVS and Nerfacto reconstructions do not contain the dining hall (bottom left in the corresponding reconstructions from \figref{fig:recon_compare}). This is because the dining hall could not be registered with the outdoor square by COLMAP (partly due to the poor lighting conditions as explained in \ref{sec:eval_loc}).

\begin{figure*}[t]
    \centering 
    \includegraphics[width=2\columnwidth]{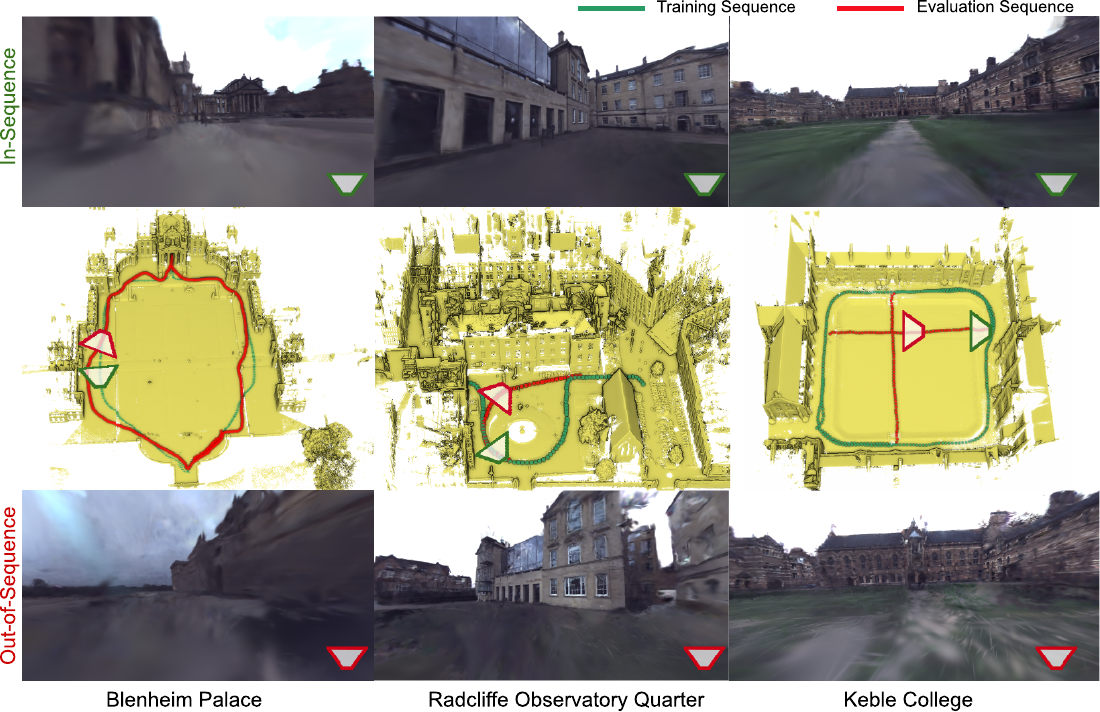}
    \caption{Illustrative results of Splatfacto-big when evaluated using in-sequence (green) and out-of-sequence (red) trajectories. When the rendering viewpoint is quite different from the training trajectory, the rendered images exhibit many more artefacts. The in-sequence and out-of-sequence trajectories in Radcliffe Observatory Quarter and Blenheim Palace are in different directions, while the trajectories in Keble College have similar viewing directions but are from distant positions. From our test, we found that Splatfacto-big generates more visual artefacts than Nerfacto-big.}
    \label{fig:nerf_compare}
\end{figure*}
\subsection{Novel View Synthesis}
\label{sec:novel_view}

We evaluate the quality of novel-view synthesis using the radiance field methods. Specifically, we evaluate:
\begin{itemize}
    \item Nerfacto~\citep{tancik2023nerfstudio} which is described in \secref{sec:3d_recon}.
    \item Splatfacto~\citep{ye2024gsplat}, an implementation of 3D Gaussian Splatting~\citep{kerbl3Dgaussians} with quality comparable to the original implementation.
\end{itemize}
We also include results using the above methods with increased representation capability, namely Nerfacto-big (Nerfacto with larger hash grid size and proposal network size, and more ray samples) and Splatfacto-big (Splatfacto with lower thresholds for densifying and culling 3D Gaussians, which results in more Gaussians being used). All methods are trained for 5000 iterations. We select one in every ten images as the in-sequence evaluation images.

\subsubsection{Evaluation Metrics:}
We measure the quality of the rendered images using the Peak Signal-to-noise Ratio (PSNR), Structural Similarity (SSIM)~\citep{wang2004ssim} and Learned Perceptual Image Patch Similarity (LPIPS)~\citep{zhang2018lpips} metrics, as commonly used in the literature~\citep{mildenhall2021nerf,barron2022mipnerf360,tancik2023nerfstudio}.

\subsubsection{Out-of-Sequence Novel View Synthesis:}
When evaluating radiance field methods, methods often use test poses that are close to the training poses. This is typically because the test poses and training poses are sampled from a common input trajectory.
In downstream applications, the ability to render photorealistic images from viewpoints that are quite different from the training poses is crucial. To facilitate research in this direction, we generate challenging test sets whose viewpoints are very different from the training sets. Specifically, we merged images from different sequences taken in the same site using COLMAP. Then, we manually selected training and test set images that are far away apart or have very different view directions. We describe the images that are selected from the input trajectory as ``in-sequence'' and images from a separate trajectory with different viewpoints as ``out-of-sequence''.

\subsubsection{Experimental Results:}

We present quantitative results in \tabref{tab:nerf_eval}. Of particular interest, one can see that the quality of novel view synthesis falls significantly when moving from the in-sequence trajectory to the out-of-sequence trajectory. Compared to Nerfacto, Splatfacto (and its big version) generalises worse in the out-of-sequence setting, and we show qualitative results of Splatfacto-big in \figref{fig:nerf_compare}. The generalisation issue is particularly evident in Radcliffe Observatory Quarter and Keble College
where the renderings are almost photorealistic from an angle close to the training data, but exhibit severe artefacts when rendered from a different location. Some of the artefacts have the wrong 3D geometry, and a typical issue is there being elongated 3D Gaussians along the training view angles as mentioned in \cite{matsuki2024gaussianslam}.
Other artefacts, such as the black artefact on the ground from Radcliffe Observatory Quarter, are due to the modelling of view-dependent colour used in radiance field methods. View-dependent colour is commonly modelled by a neural network~\citep{mildenhall2021nerf} or spherical harmonics~\citep{kerbl3Dgaussians}. When the training viewing angles are limited (which is common in robotics applications), the optimised neural network or spherical harmonics can be overfitted which leads to unexpected colours when rendering from a novel viewing angle. This is a limitation of the state-of-the-art radiance fields method which is under-explored in the literature.

\begin{figure}[h]
\centering \includegraphics[width=1.0\columnwidth]{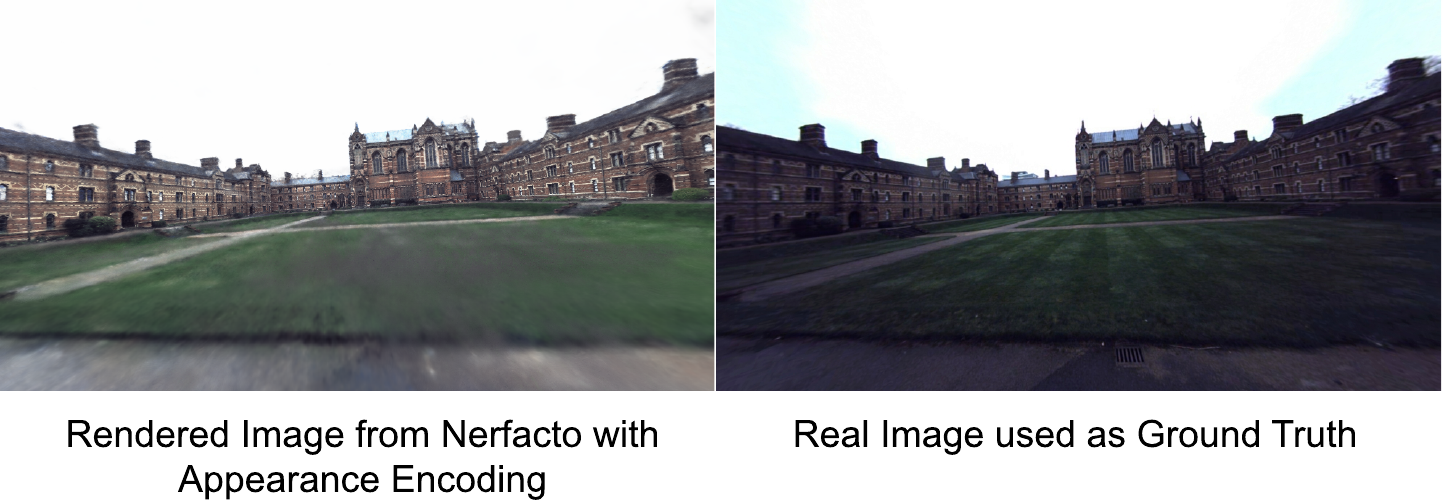}
	\caption{Comparison between an evaluation image and a rendered image from Nerfacto~\citep{tancik2023nerfstudio}. The PSNR metric is affected not only by the visual scene, but also by the lighting difference. Nerfacto uses per-frame appearance encodings~\citep{martinbrualla2020nerfw} which are optimised during training. When rendering at test time, Nerfacto uses the averaged appearance encoding of the training images. In the novel-view synthesis evaluation, we give more consideration to LPIPS and SSIM since they are more invariant to lighting differences.}
	\label{fig:render_appearance_diff}
\end{figure}

For the methods we tested, we found them all to be capable of generating reasonably photo-realistic images when rendering from in-sequence poses. A key difference between Nerfacto and Splatfacto is the rendering speed at test time: Both Splatfacto and Splatfacto-big render at 3.5 Hz on average, while Nerfacto renders at 1.25 Hz and Nerfacto-big at 0.57 Hz. When using the ``big'' version for Nerfacto and Splatfacto, the rendering quality is generally better with LPIPS increased by 9.6\% and SSIM by 2\% on average. This improvement is not always reflected in the PSNR measure, because it is also affected by the per-frame appearance difference (e.g. lighting)~\citep{martinbrualla2020nerfw} as illustrated in \figref{fig:render_appearance_diff}. For this reason, we give more consideration to changes in LPIPS and SSIM. The appearance difference issue can be potentially addressed by techniques such as test-time appearance encoding optimisation~\citep{martinbrualla2020nerfw}.

\section{Limitations and Future Work}

The AlphaSense cameras have limited dynamic range. To be able to capture both dark and bright environments, we used the auto-exposure function of the cameras during the data collection. However, images captured using auto-exposure will have inconsistent pixel intensity when observing the same 3D structure from different viewpoints. Because of this, merging colourised LiDAR point clouds would lead to a mixture of different colours in the reconstruction. We believe this is an important research question, and there are several promising directions to address this issue, including estimating image exposure time (e.g. R3LIVE++~\citep{lin2024r3live++}) and modelling image appearance as latent features (e.g. NeRF-W~\citep{martinbrualla2020nerfw}, GS-W~\citep{zhang2024gaussianw}).

Our localisation and reconstruction benchmarks primarily evaluate classical SLAM systems. The benchmarks can be extended to include more recent learning-based SLAM systems such as DROID-SLAM~\citep{teed2021droid}, Gaussian Splatting SLAM~\citep{matsuki2024monogs}, MASt3R-SLAM~\citep{murai2024mast3r} and PIN-SLAM~\citep{pan2024pinslam}. Evaluating these methods could provide more insights into their performance in large-scale outdoor data and facilitate the development of new approaches.

\section{Conclusions}
We present a large-scale dataset with colour images and LiDAR scans paired with high-quality ground truth 3D models and sensor trajectories. We demonstrate that the dataset is suitable for evaluating a variety of tasks in robotics and computer vision including LiDAR SLAM, Structure-from-Motion, Multi-View Stereo, Neural Radiance Field and 3D Gaussian Splatting. The scale of the provided data sequences and the quality of the ground truth trajectory and reconstruction make it suitable for evaluating large-scale localisation and 3D reconstruction methods in an outdoor environment. In addition, the colour cameras used in our dataset make it suitable for evaluating radiance field approaches, and encourage the development of SLAM systems integrated with radiance field representations. In particular, we demonstrate that state-of-the-art radiance field methods require further development to be applicable in the robotics context, namely inaccurate 3D geometry and limited generalisation capability when tested with poses distant from the training sequence.

\small \section*{Acknowledgements}
The authors would like to thank Tobit Flatscher for helping with the GPS sensor, Prof. Ayoung Kim, Matias Mattamala and Christina Kassab for discussions and proofreading, Haedam Oh and Jianeng Wang for helping with dataset collection and calibration, Dongjae Lee for helping with post-processing.

\small \section*{Declaration of Conflicting Interest}
The authors declared no potential conflicts of interest with respect to the research, authorship, and/or publication of this article.

\small \section*{Funding}
This project has been partly funded by the National Research Foundation of Korea (NRF) grant funded by the Korea government (MSIT)(No. RS-2024-00461409). Miguel Ángel Muñoz-Bañón is supported by the Valencian Community Government and the European Union through the CIBEST/2023/44 fellowship and PROMETEO/2021/075 project.

\bibliographystyle{SageH}
\bibliography{references.bib}

\end{document}